RESEARCH ARTICLE

# Exploring Multimodal Perception in Large Language Models Through Perceptual Strength Ratings

**JONGHYUN LEE[1,2], DOJUN PARK[3], JIWOO LEE[4,5], HOEKEON CHOI[5,6], AND SUNG-EUN LEE[3,4,5]**

[1]English Studies Major, Division of Global Studies, College of Global Business, Korea University, Sejong Campus, Sejong 30019, Republic of Korea
[2]Institute of Humanities, Seoul National University, Seoul 08826, Republic of Korea
[3]Artificial Intelligence Institute, Seoul National University, Seoul 08826, Republic of Korea
[4]Department of German Language and Literature, Seoul National University, Seoul 08826, Republic of Korea
[5]Brain and Humanities Laboratory, Seoul National University, Seoul 08826, Republic of Korea
[6]Department of English Language and Literature, Seoul National University, Seoul 08826, Republic of Korea

Corresponding author: Sung-Eun Lee (cristlo5@snu.ac.kr)

This work was supported in part by the Ascending SNU Future Leader Fellowship through Seoul National University under Grant 0524-20230015, and in part by the Basic Science Research Program through the National Research Foundation of Korea (NRF) funded by the Ministry of Education under Grant RS-2023-00274280.

**ABSTRACT** This study investigated whether multimodal large language models can achieve human-like sensory grounding by examining their ability to capture perceptual strength ratings across sensory modalities. We explored how model characteristics (size, multimodal capabilities, architectural generation) influence grounding performance, distributional factor dependencies (word frequency, embeddings, feature distances), and human-model processing differences. We evaluated 21 models from four families (GPT, Gemini, LLaMA, Qwen) using 3,611 words from the Lancaster Sensorimotor Norms through correlation, distance metrics, and qualitative analysis. Results showed that larger (6 out of 8 comparisons), multimodal (5 of 7), and newer models (5 of 8) generally outperformed their smaller, text-based, and older counterparts. Top models achieved 85-90% accuracy and 0.58-0.65 correlations with human ratings, demonstrating substantial similarity. Moreover, distributional factors showed minimal impact, not exceeding human dependency levels. However, despite strong alignment, models were not identical to humans, as even top performers showed differences in distance and correlation measures, with qualitative analysis revealing processing patterns related to absent sensory grounding. Additionally, it remains questionable whether introducing multimodality resolves this grounding deficit. Although multimodality improved performance, it seems to provide similar information as massive text rather than qualitatively different data, as benefits occurred across unrelated sensory dimensions and massive text-only models achieved comparable results. Our findings demonstrate that while advanced LLMs can approximate human sensory-linguistic associations through statistical learning, they still differ from human embodied cognition in processing mechanisms, even with multimodal integration.

## I. INTRODUCTION

One of the key criticisms of Large Language Models (LLMs) is their lack of grounding, meaning their understanding of language is not connected to real-world experiences [1], [2], [3]. Humans acquire language through a rich interplay of sensory and motor experiences, with comprehension tied to embodied experiences [4], [5], [6], [7], [8], [9], [10]. For example, understanding the word "coffee" involves not just knowing the words frequently associated with it but also our sensory

experiences—its appearance, taste, and aroma. In contrast, earlier LLMs primarily learned from vast amounts of text, resulting in language processing that remained largely disconnected from non-linguistic, real-world context. Critics argued that if human-like language understanding relies on grounding words in non-linguistic associations [11], models trained exclusively on text would never achieve human-level comprehension [1], [2], [3].

Recent attempts to integrate multimodality into LLMs [12], [13], [14] may offer a potential solution to the symbol grounding problem [15]. However, significant questions remain about whether multimodal integration has truly solved the grounding problem and, if so, to what extent it brings LLMs closer to human-like comprehension. Although these models now have access to richer contexts beyond text alone, it remains unclear how much this added information allows them to approach language understanding in a way that resembles the embodied processes seen in humans. Additionally, the performance gains observed in multimodal LLMs are not solely attributable to multimodal integration but also stem from massive increases in parameters and refinements in model architecture [16]. This raises the question of how much of the observed progress is actually driven by the incorporation of multimodal input versus more advanced text processing and inference capabilities.

To address these questions, this study evaluates multimodal LLMs using perceptual strength ratings, which measure how strongly certain words evoke sensory experiences across six modalities: visual, auditory, gustatory, olfactory, haptic, and interoceptive [17]. This approach provides a systematic framework for assessing grounding capabilities while allowing direct comparison with human performance. Building on prior work [18], [19], this study extends the evaluation through more in three ways: (1) comparing LLMs with similar architectures but differing multimodal capabilities, (2) exploring distributional factors that influence performance, and (3) conducting detailed word-level qualitative analysis to identify cases where embodied simulation may be necessary for accurate comprehension.

The research questions guiding this study are: (1) How well do multimodal LLMs capture perceptual strength ratings across various sensory modalities compared to human performance? (2) What role do multimodal capabilities, model size, and architectural improvements play in enhancing grounding performance? (3) To what extent do distributional factors influence the model performance in perceptual strength ratings? (4) Which individual words or sensory modalities reveal significant differences between human and models, and what do these differences reveal about the limitations or strengths of these models in simulating embodied experiences?

## II. RELATED WORK

### A. EMBODIED COGNITION AND LANGUAGE PROCESSING

The symbol grounding problem, originally formulated by Harnad [11], questions how symbols acquire meaning without being grounded in sensorimotor experiences. Embodied cognition theories propose one solution to this problem, suggesting that human language understanding is fundamentally grounded in sensorimotor experiences [4], [5], [6], [7], [8], [9], [10]. According to this view, comprehension involves the simulation of perceptual and motor experiences associated with linguistic concepts. For example, understanding action words like "kick" involves activating motor representations, while processing sensory words like "loud" engages auditory processing regions [20], [21], [22], [23]. However, the necessity of embodiment in language comprehension remains debated. Critics argue that embodied simulation may not be essential for language understanding, suggesting that abstract pattern recognition could be sufficient [24], [25], [26], [27].

How LLMs perform on tasks requiring embodied understanding can inform this theoretical debate. If LLMs achieve high performance through text alone, this would suggest embodiment may not be essential for language processing [24], [25], [26], [27]. If multimodal input proves necessary, this would support embodiment but suggest it may be less deeply rooted than assumed [28], [29]. If neither approach proves sufficient, this would reinforce the importance of deep embodied grounding. Each scenario has different implications for understanding both human cognition and artificial language systems.

### B. MULTIMODAL LANGUAGE MODELS

Recent advances in multimodal language models have attempted to address grounding limitations by incorporating diverse input modalities beyond text. Current multimodal architectures process combinations of text, images, audio, and video inputs [12], [13], [14], [30], representing a shift from purely text-based training approaches. As [2] metaphorically suggests, current multimodal models have expanded their scope from "reading" to "watching TV," potentially offering a solution to the symbol grounding problem [15].

These architectures typically process visual data (images and videos) alongside textual inputs, fusing different modalities to enable cross-modal understanding and text generation [31]. These models can be distinguished according to several factors such as training approaches (training from scratch [12], [32], [33], [34] *vs.* LLM-as-backbone [35], [36], [37], [38], [39], [40]) or fusion mechanisms (contrastive learning [33], [34], projection-based alignment [36], [37], [40], cross-attention mechanisms [32], [35], [41], token unification [12]).

Despite these various model structures, important questions remain about whether this multimodal integration truly resembles human embodied perception. When compared to the complex ways humans interact with their environment, this multimodal input represents a rather superficial form of embodiment. The key question is whether multimodal models can sufficiently replicate human embodied language processing with this level of superficial multimodal input.

### C. EVALUATING GROUNDING IN LANGUAGE MODELS

Previous studies have examined LLMs' grounding capabilities across various grounding-related tasks. Research has tested physical commonsense knowledge [2], [42], attribute and conceptual reasoning [43], [44], spatial and temporal understanding [44], multimodal and embodied simulation [15], [45], and problem-solving in out-of-distribution scenarios [17]. For example, [15] tested multimodal LLMs on their ability to simulate implicit visual features implied by language, such as shape, color, and orientation.

The findings from these studies are mixed: some studies reported that models struggled with more complex physical reasoning tasks [2], [42] and exhibited poor performance in out-of-distribution problem-solving scenarios [46]. In contrast, other research demonstrated that LLMs can successfully handle tasks previously thought to require grounding, such as commonsense attribute classification [12], generalizing to conceptual domains [44], and representing spatial and temporal information [47]. Overall, although the performance varies depending on the task and the model, LLMs tend to perform grounding tasks at least above chance levels, with more recent models showing greater success in addressing grounding challenges.

However, it remains unclear whether these models have reached human-level grounding or even come close, as most of the studies did not directly compare their results with human performance. Among the studies that did make such comparisons [2], [42], [43], [46], most found that LLMs still fall short of human-level performance. Furthermore, research on multimodal LLMs [15], [45] remains relatively limited.

**TABLE 1. Specifications of models used in the study.**

| Family | Name[a] | Multimodality[c] |
|---|---|---|
| GPT | GPT-3.5 | Text |
| | GPT-4o-mini | Text, Images, Audio, Video |
| | GPT-4 | Text, Images |
| | GPT-4o | Text, Images, Audio, Video |
| | GPT-4.1 | Text, Images, Audio, Video |
| Gemini | Gemini-1.5-Flash | Text, Images, Audio, Video |
| | Gemini-1.5-Pro | Text, Images, Audio, Video |
| | Gemini-2.5-Flash | Text, Images, Audio, Video |
| | Gemini-2.5-Pro | Text, Images, Audio, Video |
| LLaMA | LLaMA-3.2-3B[b] | Text |
| | LLaMA-3.2-11B-Vision[b] | Text, Images |
| | LLaMA-3.3-70B | Text |
| | LLaMA-3.2-90B-Vision | Text, Images |
| | LLaMA-3.1-405B | Text |
| | LLaMA-4-Scout-17B | Text, Images |
| | LLaMA-4-Maverick-17B | Text, Images |
| Qwen | Qwen-2.5-32B | Text |
| | Qwen-2.5-32B-VL | Text, Images, Video |
| | Qwen-2.5-72B | Text |
| | Qwen-2.5-72B-VL | Text, Images, Video |
| | Qwen-3-32B | Text |
| | Qwen-3-30B-A3B | Text |
| | Qwen-3-235B-A22B | Text |

[a] The "B" in model names indicates parameter size (e.g., "LLaMA-3.2-3B" refers to 3 billion parameters). Parameter counts for GPT and Gemini are not publicly available. GPT parameter estimates are as follows: GPT-4 (~1.76T), GPT-4o mini (~8B), GPT-4o/GPT-4.1 (~200B). Gemini models are estimated to be of similar scale. However, these are unverified public estimates and accuracy cannot be assured [63], [64]. LLaMA-4 and Qwen-3 are Mixture of Experts (MoE) models with variants: Scout and Maverick (17B active/109B-400B total, 16-128 experts), and Qwen-3-30B and -235B (3.3B-22B active/30.5B-235B total, 128 experts each). Model identifiers are provided in the supplementary materials (https://osf.io/9qgek/).
[b] Excluded from analysis due to high error response rates.
[c] Multimodality indicates each model's supported input modalities. All models were evaluated using identical text-based prompts in this study (see

### D. PERCEPTUAL STRENGTH RATINGS

Perceptual strength ratings measure how strongly words evoke sensory experiences across different modalities [17]. Initially developed to assess the embodied nature of human language processing, these ratings have been adapted into various languages, including English [17], [48], Russian [49], Dutch [50], [51], Italian [52], French [53], [54], Mandarin [55], and Korean [56]. The ratings provide insight into how people connect words with sensory experiences, revealing simulation processes that cannot be fully explained by linguistic distribution alone [57], [58]. Brain studies have validated the neurological basis of these ratings, showing that words rated for modality-specific properties align with neural activity in corresponding sensory regions [20- [23].

The application of perceptual strength ratings to evaluate LLMs is still limited. References [18] and [19] provided preliminary investigations, indicating that LLMs can approximate human-like patterns in perceptual strength ratings to some extent, though their performance remains less consistent than in other psycholinguistic dimensions such as concreteness or valence [19]. However, these early studies represent only the beginning of exploring this evaluation approach, leaving substantial room for more comprehensive research.

## III. METHODS

### A. MODELS

This study evaluated four model families comprising 23 sub-models with varying parameter sizes and multimodal capabilities, including both commercial and open-source architectures: GPT, Gemini, LLaMA, and Qwen (see Table 1 for detailed model specifications). Two models were excluded from analysis due to high error rates in response generation (detailed in the Data Trimming section), resulting in 21 models for final evaluation.

Among the open-source models, three vision-language model architectures were included: LLaMA 3.2 Vision, LLaMA-4, and Qwen-2.5-VL. LLaMA 3.2 Vision uses cross-attention adapters to integrate visual features while keeping the base language model frozen [59]. LLaMA-4 employs native multimodal training with early fusion of text and vision tokens [60]. Qwen-2.5-VL utilizes MLP-based projection to align visual embeddings with the language model's text space [61].

### B. MATERIALS

A total of 3,611 words were selected from the Lancaster Sensorimotor Norms dataset [17], which provides perceptual and action strength ratings for 40,000 English words. The selection of these 3,611 words was based on the following criteria. First, the words were chosen if the number of participants who reported knowing the word and felt capable of providing perceptual ratings (N_known.perceptual) was 20 or more, and if the percentage of participants who knew the word (Percent_known.perceptual) was at least 95 percent. These criteria ensured that the selected words were those that participants were both familiar with and confident in rating perceptually. Additionally, words identified as prepositions, pronouns, articles, or other function words by the Python NLTK part-of-speech tagger were excluded. Although perceptual strength ratings were available for these functional words, they were excluded due to their limited semantic content. Finally, when multiple words shared the same stem, only one representative word was retained using the Python NLTK Snowball stemmer.

### C. PROCEDURE

The experiment was conducted using the OpenAI chat completion API format across all platforms (OpenAI for GPT models, Google for Gemini, and HuggingFace for open-source models). For reasoning models (Gemini-2.5 and Qwen-3), reasoning mode was disabled to ensure consistent evaluation across all models without the influence of chain-of-thought processing. For each word, the models were prompted to provide perceptual strength ratings across six perceptual modalities: vision, hearing, touch, smell, taste, and interoception. The prompt given to the models was identical to the instructions provided to human participants (see Appendix A). Among the six senses, interoception, which involves the perception of internal bodily states, has been emphasized as a crucial sensory modality in understanding abstract concepts due to its role in emotion processing [64], [65], [66]. Recent studies indicate that people can effectively engage in interoceptive perception when rating words, revealing a negative correlation between interoception and vision [17], [54]. Each word was rated ten times, and the results were averaged across the iterations. The questions were presented as zero-shot prompts. After the models responded to a query, the next query was introduced without any reference to previous interactions.

### D. ANALYSIS

#### 1) DATA TRIMMING

For the analysis, trials where the models failed the evaluation for various reasons were excluded from the statistics. Such reasons included refusing to evaluate due to taboo words, responding that the sensory aspect could not be assessed, using a scale outside the intended 0-5 range, or providing more than six ratings. Two models with high error rates were excluded from further analysis: LLaMA-3-2-3B (30.67%) and LLaMA-3-2-11B-vision (10.21%). Among the remaining models included in the analysis, the overall error rate was 0.57% across all trials.

The trimmed data were averaged for each word and sensory modality, with these word-level averages being processed and utilized in various analyses.

#### 2) OVERALL PATTERNS

To introduce the overall patterns, *mean perceptual strength ratings* across the six sensory modalities for each model and human evaluations will be presented. Violin plots will be used to visualize the distribution of ratings. Three models (GPT-3.5, GPT-4, Qwen-2.5-72B) were selected for presentation alongside human ratings, representing distinct patterns observed across the full model set. Entire plots for all models and additional descriptive measures such as exclusivity scores and dominant modality distributions are provided in the online supplementary materials.

#### 3) PERFORMANCE MEASURES

To evaluate how closely each model's ratings align with human ratings, two primary performance measures were employed: correlation and distance-based metrics.

##### a: CORRELATION ANALYSIS

*Correlations* between each model and human ratings were calculated for each sensory modality using the stats.spearmanr function from the SciPy Python package [67]. To establish a reference point (human benchmark) for these correlations, inter-rater correlations among human participants were calculated by randomly splitting human ratings into two groups of 10 raters, calculating their correlation, repeating this process 1,000 times, and averaging these correlation values. Additionally, a *mean correlation* was computed by averaging correlations across all six sensory modalities to provide a single performance indicator for each model.

##### b: DISTANCE-BASED METRICS

To conduct a comprehensive comparison across all six sensory modalities simultaneously, cosine distances between model and human ratings were calculated for each word. These distances were computed in a pairwise manner, representing the similarity between the mean perceptual strengths of each word across models and human ratings. For example, the word "TEATIME" had the following mean perceptual strengths: "[auditory: 1.8, gustatory: 4.4, haptic: 2.9, interoceptive: 2.2, olfactory: 3.8, visual: 3.6]" in GPT-4o, and "[auditory: 2.0, gustatory: 2.67, haptic: 1.29, interoceptive: 1.29, olfactory: 1.81, visual: 2.71]" in human ratings. The cosine distance between these two points was calculated as 0.032 (where 0 indicates identical ratings and 1 indicates completely opposite ratings).

From these cosine distances, two metrics were derived: (1) *mean cosine distance* across all words and (2) *cosine accuracy rate*.

For the *mean cosine distance*, a human benchmark was established using human inter-rater distances, determined similarly to the inter-rater correlation. For each word, human ratings were randomly split into two groups of 10 raters, and the distance between these groups was computed. This process was repeated 1,000 times to obtain a mean inter-rater distance for each word, which was then averaged across all words to produce the overall mean inter-rater distance.

For the *cosine accuracy rate*, the same inter-rater distance permutations formed a distribution of random distances within human ratings for each word. If the model-human distance fell within the top 1 percent ($p < 0.01$) of this random distance distribution, it was considered significantly larger than expected from random human variability alone, and the model's rating was deemed inaccurate (accuracy = 0). Here, the term 'accuracy' is used for intuitive understanding rather than implying objective correctness. Since perceptual strength ratings reflect subjective human experiences of sensory associations which naturally vary across individuals, we chose the more stringent threshold of $p < 0.01$ to provide a lenient evaluation that only flags the extreme deviations from typical human perceptual patterns.

#### 4) COMPARATIVE ANALYSIS BETWEEN MODELS

To examine the factors contributing to model performance in perceptual strength ratings, systematic comparisons were conducted across three key dimensions (size, multimodality, generation) within model families, using metric-appropriate statistical methods: Fisher's Z-test (cocor [68] package in R) for modality-specific correlation comparisons; also Fisher's Z-test for average correlation comparisons, where individual correlations were first transformed to Fisher's Z scale, averaged within each model [69], then compared between models; mixed-effects linear regression (lmerTest [71 package in R, which is based on lme4 [71]) for cosine distance comparisons with models as fixed effects and words as random effects.

##### *a: MODEL SIZE: WHETHER LARGER PARAMETER COUNTS LEAD TO IMPROVED ALIGNMENT WITH HUMAN PERCEPTUAL RATINGS*

COMPARISONS:

| | **Small** | **Large** |
|---|---|---|
| GPT | GPT-4o-mini | GPT-4o |
| Gemini | Gemini-1.5-Flash | Gemini-1.5-Pro |
| | Gemini-2.5-Flash | Gemini-2.5-Pro |
| LLaMA | LLaMA-3.2-70B | LLaMA-3.1-405B |
| | LLaMA-4-Scout-17B | LLaMA-4-Maverick-17B |
| | (Total parameters: 109B *vs.* 400B) | |
| Qwen | Qwen-2.5-32B | Qwen-2.5-72B |
| | Qwen-2.5-32B-VL | Qwen-2.5-72B-VL |
| | Qwen-3-30B-A3B | Qwen-3-235B-A22B |

##### *b: MULTIMODALITY*

whether multimodal capabilities enhance performance.

COMPARISONS:

| | **Text** | **Multimodal** |
|---|---|---|
| GPT | GPT-4[a] | GPT-4o |
| | [a] Both multimodal, but GPT-4o has enhanced multimodal integration including audio and video. | |
| LLaMA | LLaMA-3.3-70B[b] | LLaMA-3.2-90B-Vision |
| | LLaMA-3.1-405B[c] | |
| | [b] Parameter size difference may confound this comparison, but this represents the closest available pairing within the LLaMA-3 family for examining multimodal effects. [c] This serves as a large-scale text-only model for comparison against smaller multimodal capabilities. | |
| Qwen | Qwen-2.5-32B | Qwen-2.5-32B-VL |
| | Qwen-3-32B[d] | |
| | Qwen-2.5-72B[e] | |
| | [d] Qwen-3-32B represents an advanced text-only model with the similar parameter size. [e] Qwen-2.5-72B represents a larger text-only baseline for comparison against the smaller multimodal model. | |
| | Qwen-2.5-72B | Qwen-2.5-72B-VL |

##### *c: ARCHITECTURAL GENERATION*

whether newer model generations demonstrate improved performance due to enhanced architecture or training methodologies. Model pair selection aimed to match similar size and multimodality.

COMPARISONS:

| | Older | | Newer |
|---|---|---|---|
| GPT | GPT-4 | GPT-4o | GPT-4.1 |
| Gemini | Gemini-1.5-Flash | | Gemini-2.5-Flash |
| | Gemini-1.5-Pro | | Gemini-2.5-Pro |
| LLaMA | LLaMA-3.2-90B-Vision | | LLaMA-4-Maverick-17B |
| Qwen | Qwen-2.5-32B | | Qwen-3-32B |
| | | | Qwen-3-30B-A3B |
| | Qwen-2.5-72B | | Qwen-3-235B-A22B |

#### 5) HUMAN-MODEL ALIGNMENT ANALYSIS

To examine how closely the models approximate human perceptual ratings, we conducted analyses comparing high-performing models against human benchmarks. Eight models were selected based on their higher accuracy (>85%): GPT-4, GPT-4o, GPT-4.1, Gemini-2.5-Flash, Gemini-2.5-Pro, LLaMA-3.2-90B-Vision, LLaMA-405B, and LLaMA-4-Maverick-17B. This analysis encompasses three dimensions: quantitative performance alignment, distributional factors, and qualitative pattern differences.

##### *a: QUANTITATIVE PERFORMANCE ALIGNMENT*

To assess the degree of alignment between top-performing models and human perceptual ratings, correlation and distance-based metrics (Section D.3) were employed.

***Correlation Analysis*:** Correlations between model and human ratings for each sensory modality were compared against human inter-rater correlation benchmarks. Permutation tests (1,000 iterations) determined whether model-human correlations significantly differed from human inter-rater performance, with models achieving non-significant differences ($p > 0.05$) considered to have reached human-level alignment in that modality.

***Distance Analysis***: Model performance was assessed by comparing mean cosine distances between each model and human ratings to the human inter-rater distance benchmark. Mixed-effects linear regression (lmerTest [70]) tested significance, with model type as a fixed effect and word as a random effect. Models achieving distances not significantly different from human inter-rater variability were considered to have reached human-level alignment.

### b: DISTRIBUTIONAL FACTOR ANALYSIS

To examine how linguistic factors influence the performance of the models in perceptual strength ratings, this study analyzed the influence of word frequency, word embeddings, and feature distances.

***Word Frequency:*** Low-frequency words may be underrepresented in LLM training data, leading to less accurate perceptual strength ratings. Unlike humans, who can rely on sensory experiences regardless of word frequency, LLMs are likely to struggle more with infrequent words. To test this, we analyze the relationship between word frequency and accuracy or distance, expecting lower accuracy or larger distance for low-frequency words. Word frequency was calculated using the zipf_frequency function from the wordfreq Python package [72]. Words were then divided into high- (mean=4.11, SD=0.61) and low-frequency (mean=2.46, SD=0.61) groups based on the median frequency. To determine whether the differences between high- and low-frequency groups are statistically significant, a linear regression model was employed, with word frequency group (high vs. low) as the fixed effect and cosine distance as the dependent variable. Additionally, the correlation between word frequency and distance was measured using Spearman's correlation (Python SciPy) to further assess the relationship between frequency and distance.

***Word Embeddings:*** In distributional semantics [73], a word's meaning is understood based on the context in which it frequently co-occurs with other words. Words that appear in similar contexts are likely to have similar sensory properties. If LLMs have learned enough contextual information, they may be able to evaluate perceptual strength ratings based solely on word distribution, without grounded sensory understanding. In this case, the similarity in word distributions would show a high correlation with perceptual strength ratings. If the model's performance is heavily influenced by distributional patterns, we would expect this correlation to be stronger than what is observed in human ratings. Word embeddings are numerical representations of words that capture their contextual relationships, with similarity between embeddings reflecting the similarity of their distributions. We calculated word embeddings using OpenAI's text-embedding-3-small model and compared cosine distances between word embeddings to distances in perceptual strength ratings using Spearman's correlation.

***Feature distance:*** Feature distance is another metric for measuring the similarity between words, based on shared concept features. This is derived by asking people to list features associated with a word (e.g., for a cat, "tail," "animal," "pet") and then calculating the similarity of these feature lists across words [74]. This method is likely to capture a more direct connection to sensory experiences than word embeddings. LLMs that have sufficiently learned these feature associations could potentially use them to generate perceptual strength ratings without direct grounding. If LLMs rely more heavily on such feature-based information compared to humans, we would expect a stronger correlation between feature distance and perceptual strength rating distances. For this analysis, feature distances were obtained from [74] for 875 overlapping words, yielding a total of 10,635 word pairs. Spearman's correlation was used to compare feature distances with perceptual strength rating distances for these word pairs.

### c: QUALITATIVE ANALYSIS OF OUTLIERS

To uncover differences between model and human evaluations that might be overlooked in broad pattern analyses, a qualitative analysis was conducted at the individual word level. The goal was to identify and investigate words where model ratings diverged significantly from human ratings. The analysis proceeded through the following stages:

Outlier Identification → Cross-Model Agreement Selection → Pattern Classification → Follow-up Analysis → Representative Example Selection

***Outlier Identification:*** Outliers were identified for each model using two criteria: (1) words labeled as incorrect based on the accuracy measure, and (2) words with residuals exceeding two standard deviations from a simple linear regression modeling the relationship between human and model ratings for each sensory modality. For criterion (2), only words where either the model or human rating was 3 or greater were included as outliers, as smaller numerical differences were insufficient to differentiate meaningful deviations between model and human ratings. These complementary criteria are designed to capture both comprehensive multi-modal divergences and specific individual sensory differences.

***Cross-Model Agreement Selection:*** After identifying outliers for each of the eight models using the established criteria, we selected words that were identified as outliers by 6 or more models (75% agreement threshold). This approach focused our analysis on words that demonstrated consistent model-human divergence across different architectures rather than model-specific anomalies. This selection process yielded 253 words for detailed qualitative analysis.

***Pattern Classification:*** These common outliers were examined in detail by the researchers and grouped into categories based on shared characteristics and underlying patterns in the divergences between model and human ratings.

***Follow-up Analysis:*** In certain instances, additional evaluations separate from the main rating process were conducted to explore the reasoning behind model assessments. In these cases, models were prompted to evaluate specific words and provide explanations for their ratings through follow-up questions. These ratings were not included in the main evaluation but were used exclusively to gain insights into potential rationales underlying the models' ratings. While we do not equate these follow-up responses with actual "thought processes" of the models, they may offer valuable insights into the reasoning patterns underlying the observed ratings.

***Representative Example Selection:*** For each identified category, typical examples were selected for detailed presentation to illustrate the broader patterns observed within each group.

## IV. RESULTS

(Complete results, figures, and statistical details can be accessed at https://osf.io/9qgek/.)

### A. OVERALL TREND

The mean perceptual strength ratings for each model and human evaluations are summarized in Table 2 and Figure 1. The human overall pattern (Figure 1-A) shows high ratings in visual modality, moderate ratings in auditory, haptic, and interoceptive modalities, and low ratings in olfactory and gustatory modalities, which reflect typical human sensory experience where vision dominates perception while smell and taste are generally less prominent. Most models resemble this basic pattern as shown in Figure 1-B (GPT-4), with slight overrating across modalities but maintaining the overall hierarchy. However, some poor-performing models such as GPT-3.5 (Figure 1-C) exhibit substantial overrating patterns that diverge from human sensory processing. These models assign considerably higher ratings across multiple modalities, particularly in a haptic sense. In contrast, a third pattern emerges with models such as Qwen-2.5-72B (Figure 1-D) that show compressed distributions across all modalities. These models frequently assign zero ratings to all sensory modalities, avoiding overrating but producing consistently lower ratings than human patterns.

### B. COMPARATIVE ANALYSIS BETWEEN MODELS

#### 1) MODEL SIZE[a]

| | Smaller | | Larger |
|---|---|---|---|
| GPT | GPT-4o-mini | < | **GPT-4o** |
| Gemini | Gemini-1.5-Flash | < | **Gemini-1.5-Pro** |
| | **Gemini-2.5-Flash** | ≥ | Gemini-2.5-Pro |
| LLaMA | LLaMA-3.3-70B | < | **LLaMA-3.1-405B** |
| | LLaMA-4-Scout-17B | < | **LLaMA-4-Maverick-17B** |
| Qwen | Qwen-2.5-32B | ≤ | **Qwen-2.5-72B** |
| | **Qwen-2.5-32B-VL** | ≥ | Qwen-2.5-72B-VL |
| | Qwen-3-30B-A3B | < | **Qwen-3-235B-A22B** |

[a] The comparison symbols indicate relative performance based on average correlation and mean cosine distance. < or >: both metrics favor the same model. ≤ or ≥: one metric favors one model, the other shows no significant difference; =: both metrics equal OR metrics contradict each other.

Larger models generally demonstrated superior alignment with human perceptual strength ratings compared to their smaller counterparts within the same family. Exceptions were Gemini-2.5-Flash outperforming Gemini-2.5-Pro on average correlation and Qwen-2.5-32B-VL outperforming Qwen-2.5-72B-VL on cosine distance. Modality-specific correlation comparisons revealed similar patterns, with larger models showing better performance in 62.5% of comparisons (Table 3). GPT and LLaMA families demonstrated consistent advantages for larger models across all sensory modalities. The size advantage for larger models was most pronounced in gustatory, haptic, and olfactory modalities, while vision showed relatively fewer advantages.

#### 2) MULTIMODALITY

| | Text | | Multimodal |
|---|---|---|---|
| GPT | **GPT-4** | > | GPT-4o |
| LLaMA | LLaMA-3.3-70B | < | **LLaMA-3.2-90B-Vision** |
| | LLaMA-3.1-405B | = | LLaMA-3.2-90B-Vision |
| Qwen | Qwen-2.5-32B | < | **Qwen-2.5-32B-VL** |
| | Qwen-3-32B | < | **Qwen-2.5-32B-VL** |
| | Qwen-2.5-72B | < | **Qwen-2.5-32B-VL** |
| | Qwen-2.5-72B | < | **Qwen-2.5-72B-VL** |

Multimodal models generally showed better performance than text-only models, with exceptions being GPT-4 *vs.* GPT-4o and LLaMA-3.1-405B *vs.* LLaMA-3.2-90B-Vision. However, it should be noted that GPT-4 is not a text-only model, as it can process image inputs. Analysis of modality-specific correlations showed that multimodal models outperformed text-only models in 61.9% of comparisons. Interoception showed the strongest multimodal advantage (100%), while vision demonstrated the weakest multimodal advantage (28.6%).

#### 3) ARCHITECTURAL GENERATION

| | Older | | Newer |
|---|---|---|---|
| GPT | **GPT-4** | > | GPT-4o |
| | GPT-4o | = | GPT-4.1 |
| Gemini | Gemini-1.5-Flash | < | **Gemini-2.5-Flash** |
| | Gemini-1.5-Pro | ≤ | **Gemini-2.5-Pro** |
| LLaMA | LLaMA-3.2-90B-Vision | < | **LLaMA-4-Maverick-17B** |
| Qwen | Qwen-2.5-32B | ≤ | **Qwen-3-32B** |
| | Qwen-2.5-32B | = | Qwen-3-30B-A3B |
| | Qwen-2.5-72B | < | **Qwen-3-235B-A22B** |

Generational comparisons revealed that newer architectures generally outperformed older models, though less consistently than size or multimodality dimensions. Within the GPT family, the oldest model (GPT-4) demonstrated the most human-like evaluations. In modality-specific correlations, newer models also showed better performance in 47.9% of comparisons, but this advantage was smaller compared to

**TABLE 2. Comprehensive performance summary: perceptual strength ratings, correlations, and distance measures across all models.**

| Model Name | | Mean Perceptual Strength Ratings | | | | | | Correlation | | | | | | | Distance | |
|---|---|---|---|---|---|---|---|---|---|---|---|---|---|---|---|---|
| | | A | G | H | I | O | V | A | G | H | I | O | V | Mean | Cosine | Acc. |
| **Human** | | **1.56** | **0.38** | **1.21** | **1.07** | **0.47** | **3.04** | **0.781** | **0.474** | **0.757** | **0.707** | **0.562** | **0.696** | **0.663** | **0.0334 (0.02)** | - |
| GPT | **3.5** | **2.99** | **1.56** | **3.97** | **2.45** | **2.59** | **3.85** | 0.535 | 0.387 | -0.098 | 0.549 | 0.257 | 0.360 | 0.332 | 0.1545 (0.08) | 37.08% (0.48) |
| | 4o-mini | 2.70 | 1.03 | 2.81 | 2.25 | 1.68 | 4.08 | 0.627 | 0.401 | 0.376 | 0.549 | 0.454 | 0.332 | 0.456 | 0.0841 (0.06) | 70.89% (0.45) |
| | **4** | **1.71** | **0.34** | **1.41** | **1.64** | **0.64** | **3.58** | 0.713 | 0.474 | 0.697 | 0.660 | 0.574 | 0.690 | 0.635 | 0.0475 (0.05) | 91.05% (0.29) |
| | 4o | 2.32 | 0.37 | 2.26 | 1.92 | 0.76 | 3.65 | 0.694 | 0.479 | 0.604 | 0.725 | 0.575 | 0.439 | 0.586 | 0.0558 (0.05) | 85.77% (0.35) |
| | 4.1 | 1.75 | 0.36 | 1.93 | 1.93 | 0.62 | 3.53 | 0.659 | 0.493 | 0.643 | 0.767 | 0.564 | 0.692 | 0.636 | 0.0576 (0.06) | 86.48% (0.34) |
| Gemini | 1.5-Flash | 0.88 | 0.23 | 0.69 | 0.57 | 0.37 | 2.23 | 0.520 | 0.322 | 0.392 | 0.463 | 0.425 | 0.541 | 0.444 | 0.1171 (0.11) | 63.34% (0.48) |
| | 1.5-Pro | 2.22 | 0.52 | 2.21 | 2.25 | 0.85 | 3.55 | 0.656 | 0.469 | 0.684 | 0.602 | 0.569 | 0.531 | 0.585 | 0.0746 (0.09) | 77.34% (0.42) |
| | 2.5-Flash | 1.99 | 0.30 | 1.54 | 1.62 | 0.56 | 3.93 | 0.709 | 0.498 | 0.577 | 0.698 | 0.576 | 0.648 | 0.618 | 0.0533 (0.06) | 88.31% (0.32) |
| | 2.5-Pro | 2.72 | 0.46 | 1.72 | 1.28 | 0.82 | 4.26 | 0.676 | 0.469 | 0.611 | 0.608 | 0.527 | 0.605 | 0.583 | 0.0532 (0.05) | 86.57% (0.34) |
| LLaMA | 3.3-70B | 0.96 | 0.21 | 1.02 | 0.22 | 0.34 | 3.63 | 0.458 | 0.390 | 0.601 | 0.284 | 0.489 | 0.514 | 0.456 | 0.1349 (0.13) | 57.86% (0.49) |
| | 3.2-90B-Vision | 1.35 | 0.22 | 1.27 | 1.23 | 0.51 | 3.98 | 0.684 | 0.408 | 0.630 | 0.699 | 0.527 | 0.638 | 0.598 | 0.0573 (0.06) | 87.75% (0.33) |
| | 3.1-405B | 1.71 | 0.29 | 1.91 | 1.27 | 0.62 | 3.80 | 0.734 | 0.479 | 0.644 | 0.673 | 0.569 | 0.624 | 0.621 | 0.0593 (0.07) | 88.10% (0.32) |
| | 4-Scout-17B | 1.47 | 0.17 | 0.84 | 0.81 | 0.32 | 2.44 | 0.498 | 0.387 | 0.458 | 0.610 | 0.461 | 0.420 | 0.472 | 0.1069 (0.11) | 67.99% (0.47) |
| | 4-Maverick-17B | 1.92 | 0.43 | 1.99 | 1.73 | 0.63 | 4.1 | 0.695 | 0.492 | 0.717 | 0.714 | 0.580 | 0.676 | 0.646 | 0.0519 (0.06) | 88.61% (0.32) |
| Qwen | 2.5-32B | 0.62 | 0.10 | 0.12 | 0.17 | 0.18 | 1.43 | 0.465 | 0.304 | 0.367 | 0.112 | 0.423 | 0.398 | 0.345 | 0.1329 (0.10) | 53.56% (0.50) |
| | 2.5-32B-VL | 2.52 | 0.88 | 2.46 | 2.07 | 1.27 | 3.67 | 0.663 | 0.465 | 0.404 | 0.638 | 0.495 | 0.403 | 0.511 | 0.0682 (0.05) | 79.09% (0.41) |
| | **2.5-72B** | **0.84** | **0.29** | **0.91** | **0.44** | **0.41** | **1.77** | 0.354 | 0.355 | 0.376 | 0.389 | 0.413 | 0.354 | 0.373 | 0.1211 (0.11) | 61.34% (0.49) |
| | 2.5-72B-VL | 2.27 | 0.73 | 2.41 | 1.68 | 1.07 | 3.38 | 0.648 | 0.429 | 0.422 | 0.559 | 0.497 | 0.333 | 0.481 | 0.0774 (0.07) | 76.85% (0.42) |
| | 3-32B | 1.42 | 0.48 | 1.46 | 2.51 | 0.75 | 3.79 | 0.508 | 0.296 | 0.317 | 0.181 | 0.448 | 0.496 | 0.374 | 0.0988 (0.07) | 59.88% (0.49) |
| | 3-30B-A3B | 1.83 | 0.76 | 2.23 | 1.94 | 0.85 | 2.63 | 0.532 | 0.369 | 0.094 | 0.562 | 0.366 | 0.200 | 0.354 | 0.1290 (0.10) | 53.30% (0.50) |
| | 3-235B-A22B | 2.34 | 0.67 | 2.73 | 2.51 | 1.12 | 3.40 | 0.596 | 0.351 | 0.406 | 0.261 | 0.421 | 0.493 | 0.421 | 0.1081 (0.08) | 57.18% (0.49) |

*Note*: A=Auditory, G=Gustatory, H=Haptic, I=Interoceptive, O=Olfactory, V=Visual. Acc. = Accuracy. Bold entries correspond to models displayed in Figure 1, while green shading indicates best performance within each model family. For human ratings, correlation and distance are computed with inter-rater ratings. Distance and accuracy values show standard deviations in parentheses, and for correlation, mean indicates average correlation across all six modalities. Full standard deviation data are available in online supplementary materials.

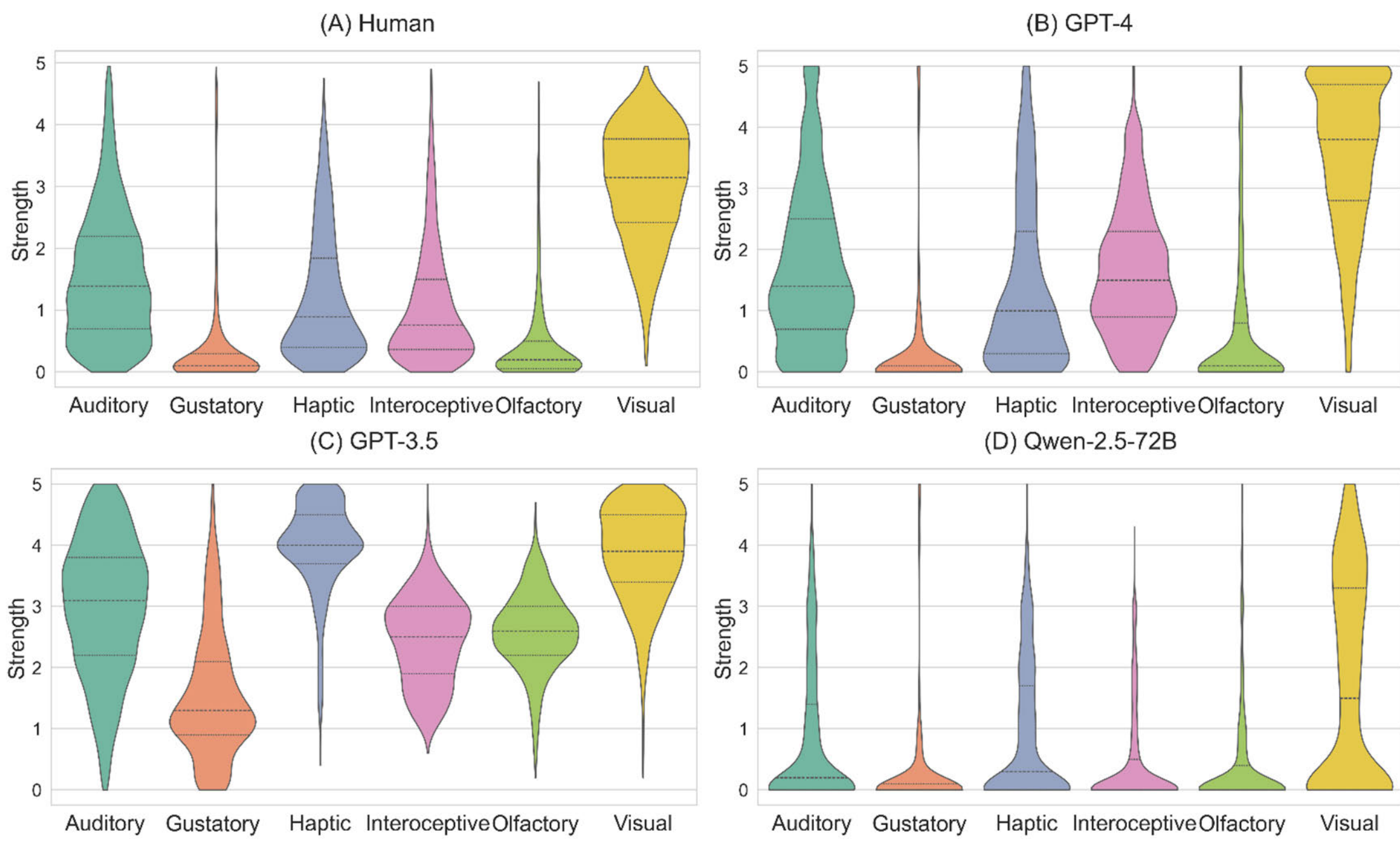

**FIGURE 1.** Violin plots illustrating the distribution of perceptual strength ratings across each sensory modality. The black lines indicate quartiles, and the width of each plot corresponds to the approximate frequency of data points at each level. Panels show: (A) Human, (B) GPT-4, (C) GPT-3.5, (D) Qwen-2.5-72B.

size or multimodal dimensions. In addition, visual modality showed the most pronounced advantage for newer models (75% of comparisons), contrasting with the weaker visual advantages observed in the other dimensions.

### C. HUMAN-MODEL ALIGNMENT ANALYSIS

#### 1) QUANTITATIVE PERFORMANCE ALIGNMENT

All selected top-performing models demonstrated significantly lower average correlations and greater mean

**TABLE 3. Model Factor performance rates in modality-specific correlations.**

| | Smaller | Larger | Text | Multi | Older | Newer |
|---|---|---|---|---|---|---|
| A | 25.0% | 62.5% | 14.3% | 71.4% | 12.5% | 50.0% |
| G | 0.0% | 62.5% | 14.3% | 57.1% | 0.0% | 37.5% |
| H | 0.0% | 75.0% | 14.3% | 42.9% | 50.0% | 37.5% |
| I | 37.5% | 62.5% | 0.0% | 100.0% | 12.5% | 62.5% |
| O | 12.5% | 62.5% | 14.3% | 71.4% | 25.0% | 25.0% |
| V | 37.5% | 50.0% | 28.6% | 28.6% | 25.0% | 75.0% |
| Total | 18.8% | 62.5% | 14.3% | 61.9% | 20.8% | 47.9% |

*Note*: A = Auditory, G = Gustatory, H = Haptic, I = Interoceptive, O = Olfactory, V = Visual. Percentages represent the proportion of significant comparisons favoring each factor within each sensory modality. Remaining percentages (not shown) represent cases with no significant difference between compared models. Size comparisons: n=8 model pairs; Multimodality comparisons: n=6 model pairs; Generation comparisons: n=8 model pairs

**TABLE 4. Correlation between representative model and human ratings across sensory modalities.**

| | | A | G | H | I | O | V | Mean |
|---|---|---|---|---|---|---|---|---|
| GPT | 4 | 0.713* | 0.474 | 0.697* | 0.660* | 0.574 | 0.690 | 0.635* |
| | 4o | 0.694* | 0.479 | 0.604* | 0.725* | 0.575* | 0.439* | 0.586* |
| | 4.1 | 0.659* | 0.493* | 0.643* | 0.767* | 0.564 | 0.692 | 0.636* |
| Gemini | 2.5-Flash | 0.709* | 0.498* | 0.577* | 0.698 | 0.576 | 0.648* | 0.618* |
| | 2.5-Pro | 0.676* | 0.469 | 0.611* | 0.608* | 0.527 | 0.605* | 0.583* |
| LLaMA | 3.2-90B-Vision | 0.684* | 0.408* | 0.630* | 0.699 | 0.527 | 0.638* | 0.598* |
| | 3.1-405B | 0.734* | 0.479 | 0.644* | 0.673* | 0.569 | 0.624* | 0.621* |
| | 4-Maverick-17B | 0.695* | 0.492* | 0.717* | 0.714 | 0.580* | 0.676* | 0.646* |
| | Models | 0.696* | 0.474 | 0.640* | 0.693* | 0.562 | 0.626* | 0.615* |
| | Human | 0.781 | 0.474 | 0.757 | 0.707 | 0.562 | 0.696 | 0.663 |

*Note*: A = Auditory, G = Gustatory, H = Haptic, I = Interoceptive, O = Olfactory, V = Visual. 'Human' represents the average inter-rater correlation among human evaluators. 'Models' represents the mean of individual model-human correlations. Asterisks (*) indicate significant differences between the model and human inter-rater correlations, determined by a permutation test (p<0.05). Colour coding indicates the direction of difference: red shades represent correlations lower than human inter-rater baseline, while blue shades represent correlations higher than human inter-rater baseline. Darker colours indicate larger magnitude differences from the human benchmark.

cosine distances compared to human inter-rater benchmarks ($p < 0.05$ for all models; Tables 4 and 5). When examining individual modalities, models consistently showed lower correlations than humans in auditory, haptic, and visual modalities. Interoceptive ratings displayed variability across models, with some exceeding human inter-rater correlations while others fell below. For gustatory and olfactory modalities, which are humans' relatively weaker senses, model performance was comparable to or slightly higher than human inter-rater correlations. To summarize, while models achieved substantial alignment with human perceptual evaluations, they did not fully reach human inter-rater benchmarks across most sensory dimensions.

### 2) LINGUISTIC DEPENDENCY ANALYSIS

***Word Frequency***: The analysis of word frequency effects revealed minimal impact on model performance. No significant differences ($p>0.05$) in cosine distance were observed between high- and low-frequency word groups for both models overall and humans. While some individual models (GPT-4o, GPT-4.1, LLaMA-4-Maverick-17B) performed worse on low-frequency words, others (Gemini models) rather exhibited the opposite pattern. Correlation analysis revealed the same trend, with both models and humans showing non-significant negligible negative correlations between word frequency and distance.

***Word Embeddings***: Correlations between word embedding distances and perceptual strength rating distances were positive across all models and humans, with all correlations around 0.15, falling within the weak range (0.1–0.3) [75]. Although some models showed slightly higher correlations than humans, suggesting a potential greater reliance on distributional semantic information, the largest gap was only about 0.02, representing a negligible difference. Several models actually showed lower correlations than humans. This indicates that while word embeddings do influence perceptual strength ratings, models do not demonstrate substantially greater dependence on this distributional cue compared to human evaluators.

***Feature distance***: Correlations between feature distances and perceptual strength rating distances were consistently positive across all models and humans, with correlation coefficients around 0.3, higher than those observed for word embeddings and reaching moderate levels (0.3-0.5) [75]. This suggests that when two words share many conceptual features, their perceptual ratings tend to be similar, reflecting the influence of semantic feature overlap on the evaluation. However, models did not show markedly greater dependence on feature similarity compared to humans. The average difference was about 0.02, and even the largest gap was around 0.05, remaining negligible. While feature similarity appears to influence perceptual ratings, this effect is not noticeably stronger in models relative to humans.

### 3) QUALITATIVE ANALYSIS OF OUTLIERS

(For words not explicitly listed under each category, refer to Appendix B. Additionally, radar charts for all outliers can be found in the online supplementary materials.)

#### *a: MODELS' MULTISENSORY OVERRATING VS. HUMAN VISION BIAS*

The majority of the identified outliers revealed a tendency for the models to overrate compared to humans. For instance, with the exception of the vision, the models assigned significantly higher ratings than humans in far more cases (Table 6).

The most prominent characteristic of models' overrating patterns was their tendency for multisensory overrating. While humans tended to strongly rate one or two sensory modalities (typically vision), models frequently exhibited a

**TABLE 5. Mean cosine distance between human ratings and each model, with frequency effects and linguistic factor correlations.**

| | | Dist. | High. | Low. | Freq. | Embed. | Feature |
|---|---|---|---|---|---|---|---|
| GPT | 4 | 0.048* | 0.047 | 0.047 | -0.007 | 0.139 | 0.343 |
| | 4o | 0.056* | 0.053 | **0.058** | -0.062 | 0.167 | 0.315 |
| | 4.1 | 0.058* | 0.055 | **0.059** | -0.035 | 0.171 | 0.328 |
| Gemini | 2.5-Flash | 0.053* | **0.056** | 0.050 | 0.047 | 0.131 | 0.300 |
| | 2.5-Pro | 0.053* | **0.056** | 0.050 | 0.045 | 0.141 | 0.290 |
| LLaMA | 3.2-90B-Vision | 0.057* | 0.059 | 0.055 | 0.013 | 0.145 | 0.327 |
| | 3.1-405B | 0.059* | 0.058 | 0.059 | 0.006 | 0.120 | 0.293 |
| | 4-Maverick-17B | 0.052* | 0.049 | **0.053** | -0.047 | 0.157 | 0.336 |
| Models | | 0.054* | 0.054 | 0.054 | -0.005 | 0.147 | 0.317 |
| Human | | 0.033 | 0.037 | 0.037 | -0.026 | 0.143 | 0.294 |

*Note*: Dist. = Mean cosine distance between model and human ratings across all words. High/Low = Mean cosine distance for high- and low-frequency word groups, respectively. Freq. = Correlation between word frequency and cosine distance. Embed = Correlation between word embedding distances and perceptual strength rating distances. Feature = Correlation between feature distances and perceptual strength rating distances. 'Human' represents the average inter-rater distance among human evaluators. 'Models' represents the mean across all individual model-human distances. Asterisks (*) indicate significant differences between model and human inter-rater distances ($p < 0.05$). Bold formatting indicates significant differences between high- and low-frequency word groups within the same model, with the bolded value representing the group with significantly larger cosine distances ($p < 0.05$). Freq, Embed, and Feature columns use color coding (red and green shading) to indicate correlation coefficients, with darker shades representing larger correlation magnitudes in either direction (positive or negative).

**TABLE 6. Count of outliers by sensory modality (deviations exceeding 2 standard deviations from the regression line).**

| Modality | A | G | H | I | O | V | Sum |
|---|---|---|---|---|---|---|---|
| Human>Model | 4 | 2 | 4 | 3 | 0 | 8 | 21 |
| Model>Human | 19 | 24 | 11 | 7 | 32 | 6 | 99 |
| Sum | 23 | 26 | 15 | 10 | 32 | 14 | 120 |

*Note*: A = Auditory, G = Gustatory, H = Haptic, I = Interoceptive, O = Olfactory, V = Visual. Human>Model indicates cases where human ratings were higher than model ratings, while Model>Human indicates the opposite. This table includes some duplicate counts. For example, the word 'beachfront' is an outlier where the model rated it significantly higher than humans in the haptic and olfactory modalities. As a result, it was counted separately for each of these modalities (haptic and olfactory).

pattern of strongly rating multiple related sensory modalities. For example, human ratings for VINEYARD showed a visually dominant pattern with a slight olfactory component, whereas the models' ratings involved multiple modalities (Figure 2a). They rated vision as strongly as humans did but also assigned high ratings to olfactory and moderate ratings to gustatory and haptic modalities. This type of multisensory evaluation accounted for nearly two-thirds of the identified outliers. Semantically, these outliers often included terms related to events, activities, jobs, places, and human body.

One intriguing aspect observed here was the models' tendency to overrate olfactory and gustatory senses (Table 6). These senses are fundamentally difficult for models to acquire through current multimodal inputs and must be inferred from linguistic information. Despite this limitation, the models did not fail to evaluate these senses due to the absence of direct sensory data; instead, they exhibit a tendency to overestimate them. For humans, olfactory and gustatory senses are generally weaker compared to other senses, leading to relatively low evaluations. In contrast, models assigned higher scores whenever linguistic cues suggested a potential association with smells or tastes.

In some respects, models seemed to capture the sensory characteristics of words more accurately than human evaluations. For example, although "vineyard" is likely a visually dominant word, it is clearly associated with olfactory qualities as well, considering elements such as the aroma of grapes, soil, and fresh air. Humans, unless they have extensive experience or knowledge related to a word, are likely to focus on one or two senses since it is inherently challenging for them to engage multiple senses simultaneously. Given the prominence of vision in human sensory perception, they are likely to evaluate words as vision-dominant. On the other hand, models can evaluate each sensory modality more independently, allowing them to incorporate multiple senses into their assessments. This may suggest models possibly surpass human capacity in capturing sensory nuances in certain cases where humans struggle with weaker senses or are unable to perform truly multisensory evaluations.

#### b: EVALUATION BASED ON LOOSE SEMANTIC ASSOCIATION

However, the models' multimodal overrating did not always result in a relatively more accurate or broader use of sensory modalities. Some of their evaluations seemed less plausible, often based on *secondary conceptual linking* or *loose semantic associations*. Instead of focusing on the sensory attributes of the word itself, they frequently included elements indirectly related to the word in its assessments. For example, in the case of VINEYARD, the models' high rating for gustatory sense was explained in a post-hoc question as follows: "Vineyards are strongly associated with the taste of grapes and wine, which are central to the vineyard experience." According to this, the gustatory connection was not directly tied to "vineyard" itself but rather to related elements such as grapes and wine. This suggests that the models' evaluation was influenced by loose semantic associations rather than a direct sensory link to the word in question.

This tendency was more pronounced in terms related to jobs and human body. For instance, the word HERBALIST was given a high olfactory rating by the models, which contrasted with human evaluations (Figure 2b). This was attributed to the word's strong association with herbs ("Herbs have strong, distinctive scents, and herbalists are strongly associated with these, making smell highly relevant."). While it is true that smelling herbs is a part of working with medicinal plants, people do not recognize a herbalist primarily through the olfactory experience of herbs, since a herbalist is a herbalist, regardless of whether herbs are physically present. Similarly, the models' evaluations of body-related terms often appeared to rely on superficial associations with sensory modalities. For example, the models strongly associated NOSTRIL with olfactory sensations (Figure 2c). Although a nostril is indeed linked to the sense of smell, people do not typically perceive the term, a nostril, itself, through olfactory experiences.

This pattern indicates that models are skilled at identifying surface-level relationships between words, possibly by their co-occurrence, and applying this understanding effectively across many words, but they might not fully comprehend deeper causal or hierarchical relationships. In their evaluations of job-related and body-related terms, the models often relied on sensory associations tied to related words (e.g., HERBALIST → herbs → olfactory) or sensory links associated with body parts (e.g., NOSTRIL → olfactory). While this strategy was effective in the case of VINEYARD, leading to a richer sensory evaluation than humans, it also resulted in less plausible evaluations in other instances.

#### c: EXPERIENCE-BASED VS. DEFINITION-BASED EVALUATION

Another characteristic revealed in the differences between model and human evaluations was that humans tend to base their assessments on personal sensory experiences or word usage, whereas models rely more on the definitional properties of language. For example, SALINE is rated highly for gustatory sense by the models but receives a lower rating from humans (Figure 2d). Saline is technically drinkable and salty by nature, so it is not implausible to rate it highly for gustatory sense based on its meaning and properties. However, since it is mostly used for medical purposes and is not typically consumed, humans did not assign it a high gustatory rating, seemingly influenced by personal experience. A similar example, but with an opposite pattern, was WHEAT (Figure 2e). Although wheat is edible, it is usually consumed in processed forms rather than in its raw state. The model assigned it a low gustatory rating, consistent with its literal meaning and properties. In contrast, humans provide high gustatory ratings for wheat, likely because they evaluated it by broadly including products made from wheat.

This pattern was also observed in disease-related terms such as LEUKEMIA (Figure 2f), PLAGUE, and PSYCHOSIS. Humans rated these words highly for interoception, likely because they evoke experiential responses such as fear or associations with internal physical discomfort. However, the models assigned lower interoceptive ratings, likely because they evaluated these terms based on the technical or medical definitions rather than the experiential responses they typically evoke in people.

These examples illustrate that humans tend to perform sensory evaluations based on personal experiences and the common usage of words, whereas models rely on dictionary definitions and inherent attributes of the words. While such models' evaluations may be considered more precise in some aspects, they are often disconnected from how people perceive and use the words in real-life contexts.

#### d: MISCELLANEOUS

***Adverbs***

Models and humans also differed in their evaluation of adverbs, where models often base their evaluation more on the adverb's base form. Humans occasionally assigned weaker ratings to these adverbs compared to their base forms. This is likely because the meaning of the base form shifts or diminishes when transformed into an adverb. For instance, with TOUGHLY (Figure 2g), humans generally rated low across all modalities, including haptic sense (while its base form ''tough'' was rated approximately 3 for haptic). It seems that ''toughly'' is used more figuratively, losing much of the haptic connotation of *tough*. However, the models rated it high in haptic sensation, and follow-up questions suggested this derived from their assessment of ''tough'' (''The term strongly aligns with the sense of touch, as 'tough' is often associated with textures'').

***Linguistic Process Terms***

Another pattern observed in outliers was that models and humans showed different modality emphasis when evaluating linguistic process terms such as COMMENT, PARAPHRSE, and SYNONYM, which relate to metalinguistic processes and communication activities. Since language involves both spoken and written forms, both humans and models evaluated these words as primarily auditory and visual. However, humans predominantly rated these words higher for auditory sense, while models consistently assigned higher ratings to the visual modality. For instance, with SYNONYM (Figure 2h), humans rated it higher for auditory sense, while models rated it higher for vision. This suggests that humans may associate these linguistic processes more with spoken language, while models may associate them more with written text processing.

## V. DISCUSSIONS

This study investigated the grounding capabilities of multimodal LLMs using perceptual strength ratings, through both quantitative overall pattern and word-level qualitative analysis. Specifically, we examined how accurately LLMs across different architectures capture human-like sensory perceptions, how multimodal integration, model size, and architectural generations influence grounding performance, how distributional factors affect model performance, and what patterns in specific words and modalities reveal about these models' capacity for embodied understanding.

### A. OVERALL PERFORMANCE PATTERNS ACROSS MODELS

The comprehensive evaluation of 21 models across four families revealed distinct performance patterns in perceptual strength rating tasks. Approximately one-third of the models achieved top-tier performance, demonstrating high accuracy rates (>85%) and strong correlations (>0.6) with human evaluations, showing strong alignment with human ratings across multiple sensory modalities. A second tier of moderate performers, comprising roughly another third of the models, though not entirely unreasonable in their evaluations, showed deviations from human sensory experiences across numerous individual words. The remaining models showed poor performance, with accuracy rates below 60% and correlations around 0.3-0.4, producing ratings that appeared largely disconnected from human sensory experiences.

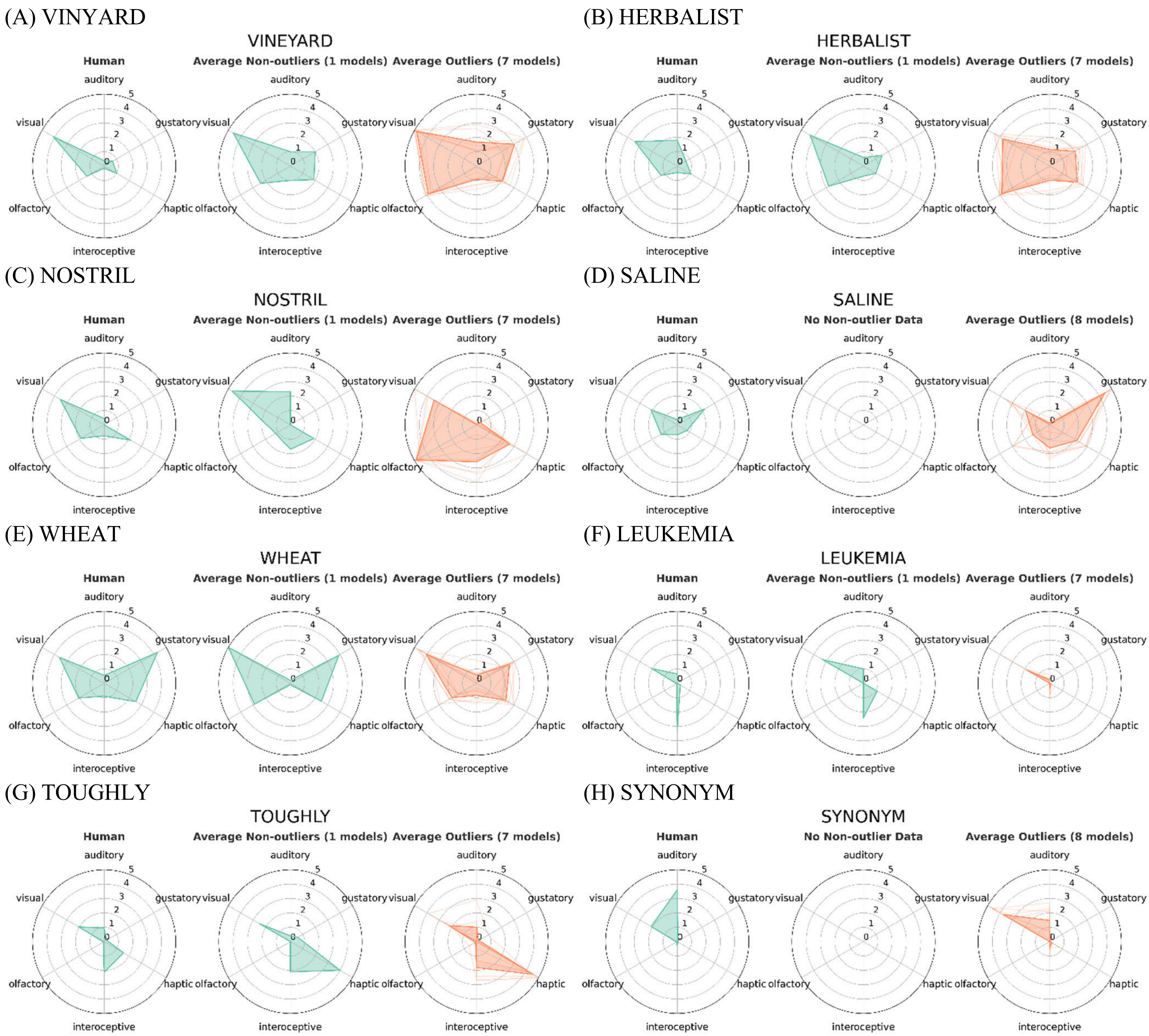

**FIGURE 2.** Selected radar charts representing average ratings across sensory modalities for outliers: (a) VINEYARD, (b) HERBALIST, (c) NOSTRIL, (d) SALINE, (e) WHEAT, (f) LEUKEMIA, (g) TOUGHLY and (h) SYNONYM. The charts depict ratings for auditory, gustatory, haptic, interoceptive, olfactory, and visual senses, arranged clockwise starting from auditory. Each word shows three panels: humas (left), non-outlier model average (center), and outlier model average (right; faint lines represent individual model evaluations). Scale ranges from 0 (center) to 5 (outer ring).

These diverse results demonstrate that the perceptual strength rating task serves as a meaningful assessment, being neither too difficult nor too easy but effectively distinguishing between models with varying capabilities. Overall, this task appeared to be a reasonably challenging task, as only one-third of the models managed to produce ratings quite similar to human perceptual patterns, while the majority of models generated evaluations that diverged from typical human sensory experiences. This might indicate that successfully capturing human-like perceptual strength ratings requires certain capabilities beyond basic language processing abilities.

This task, however, does not appear to simply measure general linguistic competence at a more challenging level. Performance patterns on perceptual strength ratings differed from those observed in other benchmarks assessing general linguistic competence or standard multimodal processing. The performance of Qwen models particularly illustrates this difference, as Qwen-2.5-VL-72B performs comparably to Llama-3.1-405B and GPT-4o on pure text tasks (MMLU-Pro [76]) and multimodal benchmarks (MMMUval [77]) respectively [61], yet showed relatively poor performance on this task. This suggests that perceptual strength ratings might probe capabilities distinct from general linguistic competence or multimodal processing. We hypothesize that performance on this task may relate to sensorimotor grounding, or how models map between sensory experiences and linguistic representations. This test could potentially measure such grounding abilities.

### B. MODEL COMPARISON: SIZE AND ARCHITECTURAL GENERATION

To understand the observed performance variations, this study compared models across three key dimensions: size, architectural generation, and multimodality. Overall, larger models, those with multimodal capabilities, and more recent architectures demonstrated superior performance.

Parameter size generally correlated with improved performance, with larger models demonstrating better alignment with human ratings. However, this relationship was not a completely positive linear correlation. When models had relatively similar parameter counts, such as Qwen-2.5 and Gemini-2.5, performance differences diminished or could even reverse. For instance, within the Qwen family, the performance gap between 32B and 72B parameter variants was minimal, suggesting that when the size difference is not substantial enough, the larger model does not necessarily outperform the smaller one.

Newer model generations generally outperformed their predecessors, but this advantage was not consistent across all cases. For example, both Qwen-3-32B and Qwen-3-30B-A3B failed to demonstrate markedly superior performance to their Qwen-2.5 counterparts. This may be attributed to their relatively lightweight nature at 30B parameters. Architectural development alone may not be sufficient to overcome size limitations, suggesting that this task may require substantial computational capacity that only larger models can provide.

Also, GPT-4, despite being an older generation, achieved the highest performance within its family, surpassing both GPT-4o and GPT-4.1. Without access to detailed architectural specifications, we can only speculate about why GPT-4 outperformed its newer counterparts. It appears that once models reach a certain performance level, further improvements become difficult to achieve, with accuracy rates of 85-90% representing near-optimal performance in this task. Another possibility is that GPT-4 might be a larger model that performs more intensive and broader computations during inference, as suggested by its higher operational costs, if we assume that the cost reflects computational complexity rather than mere inefficiency.

One additional aspect of architectural development worth examining is the performance of MoE models. Our evaluation included several MoE architectures—two Qwen-3 variants and two LLaMA-4 models—which revealed both the potential and limitations of this approach. Some MoE models showed efficiency advantages. For example, LLaMA-4-Maverick-17B achieved nearly the best performance among all LLaMA models using only 17B active parameters. Similarly, Qwen-3-30B-A3B achieved performance comparable to Qwen-2.5-32B only with 3B active parameters. These results suggest that MoE models may achieve performance levels that reflect their total parameter capacity despite using only a fraction actively, as if the broader parameter pool contributes even when not directly engaged.

However, this pattern was not consistently observed across all MoE models, as demonstrated by LLaMA-4-Scout-17B and Qwen-3-235B-A22B. In contrast to the other MoE models, they failed to achieve performance levels matching their total parameter counts. For instance, LLaMA-4-Scout, with 109B total parameters, significantly underperformed compared to LLaMA-3.2-90B-Vision. Since it has identical active parameters to the successful Maverick variant, the issue is not with active parameter count but may relate to expert configuration, as Scout uses 16 experts while Maverick employs 128. Regardless of the specific underlying factors, the success of MoE architectures in this task seems to depend on complex factors including expert selection mechanisms, routing efficiency, and task-specific optimization, rather than simple parameter scaling relationships.

### C. MODEL COMPARISON: MULTIMODALITY EFFECTS

A central focus of this study was examining multimodality effects on perceptual strength rating performance. We found that models with multimodal capabilities generally demonstrated superior performance compared to text-only counterparts. In all cases within the Qwen family, multimodal versions surpassed text-only models, with Qwen-2.5-32B-VL even exceeding both newer generation models and larger 72B variants. Also, within the LLaMA family, LLaMA-3.2-90B-Vision outperformed the similar-sized LLaMA-3.3-70B text model and achieved performance levels matching the much larger LLaMA-3.1-405B. These results indicate that multimodal models generally exceeded their text-only counterparts of comparable size and architecture, and in some cases even overcame parameter size differences.

One intriguing aspect of these findings is that although this task was text-based with no multimodal input present, multimodal training still provided benefits. In contrast, on other text benchmarks such as MMLU, there were no differences between Qwen-2.5-72B-VL and LLaMA-3.2-90B-Vision and their text counterparts [61], [78]. This is unsurprising since these vision models were built by adding vision encoders and multimodal training to their base LMs, with the additions being primarily multimodal processing capabilities. Nevertheless, in this task, multimodal models showed advantages. This suggests that multimodality may not simply enhance general linguistic capacity, but rather improve models' ability to connect word concepts with sensory experiences. This divergent pattern might simply reflect the greater difficulty of this task over other benchmarks or the difference in evaluation methodology, which measures human-like similarity rather than binary correctness. However, even if that is the case, multimodality at least enables better performance on more difficult tasks or more human-like responses.

This multimodal advantage, however, does not arise from modality-specific enhancements, contrary to what might be expected if multimodal training operated as if new sensory organs were added to a human. One might intuitively predict that the addition of a visual channel would enhance visual perception and an additional auditory channel would improve auditory perception. Yet, our modality-specific results do not support this expectation. For instance, GPT-4o's enhanced auditory input capabilities compared to GPT-4 did not result in more human-like auditory evaluations. Moreover, across all multimodality comparisons, visual modality showed the smallest improvement despite being the most prevalent training modality, while interoception, associated with abstract concepts and emotional processing [64], [65], [66] and seemingly most unrelated to multimodal inputs, showed the largest gains. These results indicate that the modality of the input itself may not directly enhance evaluation abilities for that same modality.

Indeed, the assumption that specific sensory modalities would exclusively enhance corresponding sensory perceptions may be overly simplistic. Visual input, for instance, conveys not only direct information about visible features but also indirect cues about non-visible properties, often through absence. Prior findings indicate negative correlations between vision and interoception [17], [54], implying that interoceptive states may be inferred in part from the lack of visual evidence. Moreover, visual input may also convey information relevant to other sensory modalities. For instance, an image of a person eating 'caviar' not only presents visual details but may also convey information about 'eating' caviar or its gustatory properties. In this respect, the addition of a particular modality might not operate in isolation but may instead provide comprehensive sensory information. This could help explain why improvements are observed even in sensory dimensions that appear unrelated to the added modality.

However, this form of sensory information acquisition through multimodal input still constitutes indirect rather than direct sensory experience. If we frame direct sensory experience as tasting 'caviar' firsthand and indirect experience as inferring taste associations through visual observation, then current results indicate that models improved their understanding of sensory-linguistic connections through this indirect pathway. This indirect sensory information may not provide qualitatively different dimensions of knowledge compared to what can be acquired through text alone. Indeed, our findings showed that massive text-only models such as LLaMA-3.1-405B can achieve performance levels matching other high-performing multimodal models, suggesting that extensive text data can also convey similar sensory-associative information. Text descriptions of people eating caviar, for instance, might provide comparable indirect sensory associations to visual observations of the same activity.

From the model's perspective, such integration of the different inputs might not represent the merging of truly distinct modalities. While humans perceive text and images as separate modalities, for the model, models may process both as binary data (zeros and ones) that differ only in arrangement or dimensionality. Thus, combining linguistic information with its corresponding image data might merely result in additional numerical input rather than an approximation of a grounded sensory experience. Such computational integration seems different from the human process of grounding language in sensory experiences, making it difficult to conclude that the advantages brought by multimodality stem from truly sensory grounding.

Instead, multimodality may serve to convey the same sensory information that massive text datasets can deliver, but more efficiently through more information-dense inputs [79], [80]. Admittedly, from a training perspective, smaller multimodal models may consume more resources than larger text-only models due to complex input processing and expanded training requirements. Nevertheless, the smaller parameter requirements offer the potential for reduced inference computational demands [81], [82]. In this regard, multimodal input may offer efficiency gains for improving models' sensory-linguistic connections.

### D. HUMAN-MODEL ALIGNMENT AND DIVERGENCE PATTERNS

To examine how closely models approximate human perceptual ratings, we conducted analyses comparing eight top-performing models (accuracy>85%) against human benchmarks. These models represented three major families: GPT, Gemini, and LLaMA. With the exception of LLaMA-3.1-405B, all evaluated models were multimodal architectures.

These top-performing models demonstrated substantial alignment with human perceptual strength ratings, achieving low cosine distances (0.048-0.059) and strong correlations (0.58-0.65) with human evaluations. Additionally, when examining individual modalities, the models showed correlation levels comparable to human inter-rater benchmarks in olfactory and gustatory senses and high correlations approaching 0.6-0.7 in other modalities. This level of performance is consistent with previous findings [18], [19] on LLMs' ability to approximate human-like perceptual patterns. These results suggest that they can simulate aspects of human embodied language without direct sensory experience, whether through increased size, multimodality, architectural improvements, or other factors.

However, it should be noted that perceptual strength ratings are not purely embodiment-dependent, even for humans [56], [57]. According to [58], co-occurrence frequency alone enabled differentiation among three of the five senses in corpus analysis. This indicates that perceptual strength ratings could partially be performed using purely semantic distributional information. For instance, if a model learns that words like "coffee" frequently co-occur with "aroma" or "drink," it could infer olfactory or gustatory associations. These findings suggest that models' advanced semantic association capabilities alone, whether acquired through text or multimodal inputs, may be sufficient for achieving reasonable performance on perceptual strength ratings tasks, without requiring direct sensory experience.

Given this perspective, we hypothesized that if models rely primarily on distributional information, this should manifest as greater distributional dependency compared to humans. To test this, we examined word frequency, word embeddings, and feature distance. However, the results revealed minimal evidence that such reliance substantially influenced performance. While some individual models showed slight frequency effects, consistent trends were not observed across other factors. The models displayed practically negligible differences in how word embeddings and feature distances influence perceptual strength ratings. Overall, although distributional factors influenced models' evaluations to some extent, these effects were not greater in models than in humans.

Despite this strong overall alignment, models' evaluations were not identical to human benchmarks. All top-performing models demonstrated significantly larger cosine distances and smaller average correlations compared to human inter-rater benchmarks. For individual sensory modalities, the models consistently achieved lower correlations than the benchmark in auditory, haptic, and visual modalities. Although they showed no significant differences in gustatory and olfactory modalities, this alignment likely stems from these being weaker senses for humans, resulting in greater human inconsistency. Indeed, they exhibited relatively low correlation coefficients for these two senses and patterns of frequently overrating both.

The differences became more pronounced in word-level qualitative analysis, revealing several divergence patterns. These include the models' tendency for multisensory overrating, often assigning high ratings across multiple sensory modalities compared to humans' more focused evaluations; the overestimation of olfactory and gustatory modalities, based on semantic cues rather than embodied experience; reliance on loose semantic associations, leading to sensory ratings influenced by indirectly related concepts; reliance on definitional properties, resulting in evaluations that diverge from human experiential understanding.

These differences collectively highlight that models process and understand the world in ways that are distinct from human perception to some extent. Such evaluations are, however, not always incorrect or implausible. Rather, they might be considered more comprehensive responses, while different from human interpretations, or even what could be called 'superhuman'-like responses. For instance, models appeared capable of effortlessly capturing sensory associations that 'ordinary' humans are bound to overlook due to perceptual limitations, such as the olfactory sense of 'vineyard.' They are able to conduct such multifaceted evaluations likely because they have access to information derived from vastly larger datasets than those available to humans. These characteristics may allow for the immediate retrieval and utilization of humanity's accumulated knowledge—stored in textual and other forms—to interpret and engage with the world in novel ways, serving as valuable tools for overcoming human sensory limitations.

Nevertheless, since models relied on indirect information, their evaluations could deviate from the real physical world, occasionally leading to errors. For instance, sensory evaluations based on loose semantic associations produced implausible conclusions against human experiential understanding, such as treating "NOSTRIL" as perceivable through smell. Similarly, an overreliance on definitional properties resulted in interpretations unsuitable for humans, such as connecting "SALINE" with taste despite it not being actually consumed. This pattern suggests that models may excel at identifying surface-level word relationships but, if they rely only on indirect information, these relationships might not always reflect actual sensory causality or real-world usage patterns.

### E. IMPLICATIONS FOR EMBODIED COGNITION THEORY

The strong performance of LLMs on perceptual strength rating tasks presents a challenge to strong embodiment theories [4], [5], [6], [7], [8], [9], [10]. Our findings demonstrate that multiple models achieved high accuracy rates and strong correlations with human evaluations across sensory modalities. This suggests that statistical learning from large-scale textual and multimodal data can approximate many aspects of embodied language processing without requiring the direct sensorimotor grounding. However, this challenge to strong embodiment is not absolute. The persistent qualitative differences we observed indicate that statistical approximation, while impressive, does not fully replicate human embodied cognition. These divergences suggest that direct sensorimotor experience contributes qualitatively different information that cannot be entirely captured through distributional learning alone. Therefore, our results align with graded or hybrid models of embodiment [28], [29], rather than supporting either purely embodied or purely symbolic accounts of cognition. In other words, human cognition may benefit from direct sensorimotor grounding while also utilizing abstract pattern recognition capabilities similar to those shown by LLMs.

## VI. CONCLUSION

This study investigated the grounding capabilities of multimodal language models. Our findings demonstrated that advanced LLMs closely approximate human perceptual strength ratings across sensory modalities, achieving high accuracy and strong correlations with human evaluations. However, qualitative analysis revealed systematic differences in processing mechanisms, including multisensory overrating and reliance on loose semantic associations. Furthermore, while multimodal capabilities enhanced performance, they appear to provide similar sensory information to text but more efficiently, rather than providing direct sensory grounding.

From a practical AI design perspective, this finding suggests that merely adding multimodal inputs, regardless of which sensory modalities, may not guarantee genuine sensory grounding. This points toward architectural innovations that better reflect human sensory grounding systems. Potential directions include embodied AI systems that ground language learning through physical exploration and integrate sensorimotor loops enabling real-time connections between sensory information and language processing [83], [84], [85]. Either way, at its core, architecture needs to be designed where distinct sensory modalities remain separate yet become dynamically connected to language processing as necessary, similar to how humans engage sensory simulation during language comprehension.

While our findings provide valuable insights into embodied cognition in language understanding, they were limited by several factors. First, the proprietary nature of many state-of-the-art models prevented fully controlled comparisons, though we expanded our evaluation to include open-source models with greater transparency. Second, our qualitative analysis was based on a relatively small subset of outliers and involves inherent subjectivity in pattern identification, despite efforts to employ multiple researchers and cross-model consensus. Third, our evaluation was limited to a single task focused on perceptual strength ratings, which may not encompass the full range of grounding capabilities required for comprehensive assessment. Finally, the human benchmark dataset represents rapid evaluations from approximately 20 participants per word, which may not fully capture the breadth of human perceptual experiences or deeper processing conditions.

Future research could address these limitations through several complementary approaches. First, controlled experiments with models trained under specified conditions could help overcome architectural transparency issues by enabling systematic isolation of architectural factors and multimodal integration effects. Second, future studies could benefit from comprehensive category validation using larger, systematically selected word sets across diverse semantic domains, such as medical terminology and culinary vocabulary. Finally, developing comprehensive evaluation frameworks that test multiple aspects of embodied cognition such as spatial reasoning and motor simulation, across larger, more diverse participant groups could move beyond single-task evaluation based on small-scale human benchmarks.

## APPENDIX A
## THE PROMPT GIVEN TO THE MODELS

You will be asked to rate how much you experience everyday concepts using six different perceptual senses. There are no right or wrong answers so please use your own judgement.

The rating scale runs from 0 (not experienced at all with that sense) to 5 (experienced greatly with that sense). Assign a number for each sense (Provide only ratings without explaining why).

To what extent do you experience, ‘‘WORD’’
By sensations inside your body
By tasting
By smelling
By feeling through touch
By hearing
By seeing

## APPENDIX B
## EXTENDED WORD LIST FOR EVALUATION CATEGORIES

### A. MODEL'S MULTISENSORY OVERRATING VS. HUMAN VISION BIAS

Location-related Terms

BACKYARD, BATHHOUSE, BEACHFRONT, COASTLINE, CONCERT, CREAMERY, CUBICLE, GARDEN, HOME, HOMETOWN, HOUSEWARMING, KINDERGARTEN, LAKEFRONT, LAKESIDE, MOUNTAIN, NIGHTCLUB, PINEWOOD, RIPTIDE, SMOKEHOUSE, SWAMPY, TOILETRY, VINEYARD,
WOODLAND, WOODSHOP

Job-related Terms

GLUTTON, HERBALIST, INTERN, KICKER,
LITTERBUG, LUMBERJACK, LUNATIC

Body-related Terms

BREAKDANCE, CHEW, EARPIECE, GENITAL,
GETUP, HEAVYWEIGHT, NIBBLE, NOSTRIL, PELVIC, PHYSIQUE, POSITION, QUENCHABLE, SHUDDER, SKI, SNIFF, SNOT, SNOTTY, SOFTBALL, SWALLOW, TAEKWONDO,
TONGUE, TREADMILL

Event-related Terms

AUTUMN, BONFIRE, CRISP, DINE, HANGOUT, HOLIDAY, NIGHTFALL, NIGHTTIME, ONSTAGE, RECESS, SHOWER, STORMILY, SUBMERSED, SUMMER,
TEATIME, THUNDERSTORM

Olfactory & Gustatory Overemphasis

COKE, COLA, DELI, DIET, DISINFECT, FATTINESS, FIZZY, FROTHY, HEARTBURN, HOMEMADE, JUICE, LITTERBUG, MERRINESS, NECTAR, PINEWOOD, SELTZER, SOAPINESS, TASTER, TASTINESS, TOFFEE, UPHOLSTERY, VINE

### B. EXPERIENCE-BASED EVALUATION VS. DEFINITION-BASED EVALUATION

COKE, COLA, DIBS, FIERY, FLAVOR, GAMEY, INCONSUMABLE, LEUKEMIA, PENICILLIN, PET, PUG, PUN-

GENCY, SALINE, SELTZER, SHUTTER, SON, SPERM, THING, WHEAT

### *C. MISCELLENEOUS*

Adverbs

CALMINGLY, DISGUSTEDLY, DISTRACTEDLY, DRUNKENLY, EXHILARATINGLY, FRESHLY, RESTRICTEDLY, STORMILY, TOUGHLY, UNYIELDINGLY

Linguistic Process Terms

COMMENT, COMPLAINT, MIDSENTENCE, MISPRONUNCIATION, PARAPHRASE, PHRASEMAKING, PROPOSITION, RENAME, RETOLD, SYNONYM

## REFERENCES

[1] E. M. Bender and A. Koller, "Climbing towards NLU: On meaning, form, and understanding in the age of data," in *Proc. 58th Annu. Meeting Assoc. Comput. Linguistics*, 2020, pp. 5185–5198, doi: 10.18653/v1/2020.acl-main.463.

[2] Y. Bisk, R. Zellers, R. L. Bras, J. Gao, and Y. Choi, "PIQA: Reasoning about physical commonsense in natural language," in *Proc. AAAI Conf. Artif. Intell.*, 2020, vol. 34, no. 5, pp. 7432–7439, doi: 10.1609/aaai.v34i05.6239.

[3] R. Tamari, C. Shani, T. Hope, M. R. L. Petruck, O. Abend, and D. Shahaf, "Language (Re)modelling: Towards embodied language understanding," in *Proc. 58th Annu. Meeting Assoc. Comput. Linguistics*, 2020, pp. 6268–6281, doi: 10.18653/v1/2020.acl-main.559.

[4] L. W. Barsalou, "Perceptual symbol systems," *Behav. Brain Sci.*, vol. 22, no. 4, pp. 577–660, Aug. 1999, doi: 10.1017/s0140525x99002149.

[5] L. W. Barsalou, "Grounded cognition," *Annu. Rev. Psychol.*, vol. 59, no. 1, pp. 617–645, Jan. 2008, doi: 10.1146/annurev.psych.59.103006.093639.

[6] B. K. Bergen, *Louder Than Words: The New Science of How The Mind Makes Meaning*. New York, NY, USA: Basic Books, 2012.

[7] B. Bergen and J. Feldman, "Embodied concept learning," in *Handbook of Cognitive Science*, 2008, pp. 313–331.

[8] A. M. Glenberg and M. P. Kaschak, "Grounding language in action," *Psychonomic Bull. Rev.*, vol. 9, no. 3, pp. 558–565, Sep. 2002, doi: 10.3758/bf03196313.

[9] F. Pulvermüller, "Words in the brain's language," *Behav. Brain Sci.*, vol. 22, no. 2, pp. 253–279, Apr. 1999, doi: 10.1017/s0140525x9900182x.

[10] B. Winter and B. Bergen, "Language comprehenders represent object distance both visually and auditorily," *Lang. Cognition*, vol. 4, no. 1, pp. 1–16, Mar. 2012, doi: 10.1515/langcog-2012-0001.

[11] S. Harnad, "The symbol grounding problem," *Phys. D, Nonlinear Phenomena*, vol. 42, nos. 1–3, pp. 335–346, Jun. 1990, doi: 10.1016/0167-2789(90)90087-6.

[12] D. Driess et al., "PaLM-E: An embodied multimodal language model," in *Proc. 40th Int. Conf. Mach. Learn.*, 2023, pp. 8469–8488.

[13] R. Girdhar, A. El-Nouby, Z. Liu, M. Singh, K. V. Alwala, A. Joulin, and I. Misra, "ImageBind one embedding space to bind them all," in *Proc. IEEE/CVF Conf. Comput. Vis. Pattern Recognit. (CVPR)*, Jun. 2023, pp. 15180–15190, doi: 10.1109/CVPR52729.2023.01457.

[14] S. Huang, D. Li, W. Wang, Y. Hao, S. Singhal, S. Ma, T. Lv, L. Cui, O. K. Mohammed, Q. Liu, K. Aggarwal, Z. Chi, J. Björck, V. Chaudhary, S. Som, S. Xia, and F. Wei, "Language is not all you need: Aligning perception with language models," in *pROC. Adv. Neural Inf. Process. Syst.*, 2023, pp. 72096–72109.

[15] C. R. Jones and S. Trott, "Multimodal language models show evidence of embodied simulation," in *Proc. 2024 Joint Int. Conf. Comput. Linguistics, Lang. Resour. Eval.*, 2024, pp. 11928–11933.

[16] J. Kaplan, S. McCandlish, T. Henighan, T. B. Brown, B. Chess, R. Child, S. Gray, A. Radford, J. Wu, and D. Amodei, "Scaling laws for neural language models," 2020, *arXiv:2001.08361*.

[17] D. Lynott, L. Connell, M. Brysbaert, J. Brand, and J. Carney, "The lancaster sensorimotor norms: Multidimensional measures of perceptual and action strength for 40,000 English words," *Behav. Res. Methods*, vol. 52, no. 3, pp. 1271–1291, Jun. 2020, doi: 10.3758/s13428-019-01316-z.

[18] S. Trott, "Can large language models help augment English psycholinguistic datasets?" *Behav. Res. Methods*, vol. 56, no. 6, pp. 6082–6100, Jan. 2024, doi: 10.3758/s13428-024-02337-z.

[19] Q. Xu, Y. Peng, S. A. Nastase, M. Chodorow, M. Wu, and P. Li, "Does conceptual representation require embodiment? Insights from large language models," 2023, *arXiv:2305.19103*.

[20] M. Harpaintner, E.-J. Sim, N. M. Trumpp, M. Ulrich, and M. Kiefer, "The grounding of abstract concepts in the motor and visual system: An fMRI study," *Cortex*, vol. 124, pp. 1–22, Mar. 2020, doi: 10.1016/j.cortex.2019.10.014.

[21] M. Harpaintner, N. M. Trumpp, and M. Kiefer, "Time course of brain activity during the processing of motor- and vision-related abstract concepts: Flexibility and task dependency," *Psychol. Res.*, vol. 86, no. 8, pp. 2560–2582, Nov. 2022, doi: 10.1007/s00426-020-01374-5.

[22] M. Kiefer, E.-J. Sim, B. Herrnberger, J. Grothe, and K. Hoenig, "The sound of concepts: Four markers for a link between auditory and conceptual brain systems," *J. Neurosci.*, vol. 28, no. 47, pp. 12224–12230, Nov. 2008, doi: 10.1523/jneurosci.3579-08.2008.

[23] N. M. Trumpp, F. Traub, F. Pulvermüller, and M. Kiefer, "Unconscious automatic brain activation of acoustic and action-related conceptual features during masked repetition priming," *J. Cognit. Neurosci.*, vol. 26, no. 2, pp. 352–364, Feb. 2014, doi: 10.1162/jocn_a_00473.

[24] B. Z. Mahon and A. Caramazza, "A critical look at the embodied cognition hypothesis and a new proposal for grounding conceptual content," *J. Physiol.-Paris*, vol. 102, nos. 1–3, pp. 59–70, Jan. 2008, doi: 10.1016/j.jphysparis.2008.03.004.

[25] J. R. Binder, C. F. Westbury, K. A. McKiernan, E. T. Possing, and D. A. Medler, "Distinct brain systems for processing concrete and abstract concepts," *J. Cognit. Neurosci.*, vol. 17, no. 6, pp. 905–917, Jun. 2005, doi: 10.1162/0898929054021102.

[26] R. Dalla Volta, M. Fabbri-Destro, M. Gentilucci, and P. Avanzini, "Spatiotemporal dynamics during processing of abstract and concrete verbs: An ERP study," *Neuropsychologia*, vol. 61, pp. 163–174, Aug. 2014, doi: 10.1016/j.neuropsychologia.2014.06.019.

[27] J. Wang, J. A. Conder, D. N. Blitzer, and S. V. Shinkareva, "Neural representation of abstract and concrete concepts: A meta-analysis of neuroimaging studies," *Human Brain Mapping*, vol. 31, no. 10, pp. 1459–1468, Oct. 2010, doi: 10.1002/hbm.20950.

[28] S. Danziger, J. Levav, and L. Avnaim-Pesso, "Extraneous factors in judicial decisions," *Proc. Nat. Acad. Sci. USA*, vol. 108, no. 17, pp. 6889–6892, Apr. 2011, doi: 10.1073/pnas.1018033108.

[29] D. R. Proffitt, M. Bhalla, R. Gossweiler, and J. Midgett, "Perceiving geographical slant," *Psychonomic Bull. Rev.*, vol. 2, no. 4, pp. 409–428, Dec. 1995, doi: 10.3758/bf03210980.

[30] OpenAI. (2024). *Hello GPT-4o*. Accessed: Feb. 24, 2025. [Online]. Available: https://openai.com/index/hello-gpt-4o/

[31] Z. Li, X. Wu, H. Du, F. Liu, H. Nghiem, and G. Shi, "A survey of state of the art large vision language models: Alignment, benchmark, evaluations and challenges," 2025, *arXiv:2501.02189*.

[32] L. Harold Li, M. Yatskar, D. Yin, C.-J. Hsieh, and K.-W. Chang, "VisualBERT: A simple and performant baseline for vision and language," 2019, *arXiv:1908.03557*.

[33] J. Li, D. Li, C. Xiong, and S. C. H. Hoi, "BLIP: Bootstrapping language-image pre-training for unified vision-language understanding and generation," in *Proc. Int. Conf. Mach. Learn.*, 2022, pp. 12888–12900.

[34] A. Radford, J. W. Kim, C. Hallacy, A. Ramesh, G. Goh, S. Agarwal, G. Sastry, A. Askell, P. Mishkin, J. Clark, G. Krueger, and I. Sutskever, "Learning transferable visual models from natural language supervision," in *Proc. 38th Int. Conf. Mach. Learn.*, 2021, pp. 8748–8763.

[35] J.-B. Alayrac et al., "Flamingo: A visual language model for few-shot learning," in *Proc. Adv. Neural Inf. Process. Syst.*, vol. 35, 2022, pp. 23716–23736.

[36] P. Wang, S. Bai, S. Tan, S. Wang, Z. Fan, J. Bai, K. Chen, X. Liu, J. Wang, W. Ge, Y. Fan, K. Dang, M. Du, X. Ren, R. Men, D. Liu, C. Zhou, J. Zhou, and J. Lin, "Qwen2-VL: Enhancing vision-language model's perception of the world at any resolution," 2024, *arXiv:2409.12191*.

[37] P. Agrawal et al., "Pixtral 12B," 2024, *arXiv:2410.07073*.

[38] A. Grattafiori et al., "The Llama 3 herd of models," 2024, *arXiv:2407.21783*.

[39] Z. Wu et al., "DeepSeek-VL2: Mixture-of-experts vision-language models for advanced multimodal understanding," 2024, *arXiv:2412.10302*.

[40] W. Dai, N. Lee, B. Wang, Z. Yang, Z. Liu, J. Barker, T. Rintamaki, M. Shoeybi, B. Catanzaro, and W. Ping, ''NVLM: Open frontier-class multimodal LLMs,'' 2024, *arXiv:2409.11402*.

[41] H. Lin, X. Cheng, X. Wu, and D. Shen, ''CAT: Cross attention in vision transformer,'' in *Proc. IEEE Int. Conf. Multimedia Expo (ICME)*, Jul. 2022, pp. 1–6.

[42] M. Forbes, A. Holtzman, and Y. Choi, ''Do neural language representations learn physical commonsense?'' 2019, *arXiv:1908.02899*.

[43] J. Da and J. Kasai, ''Understanding commonsense inference aptitude of deep contextual representations,'' in *Proc. 1st Workshop Commonsense Inference Natural Lang. Process.*, 2019, pp. 1–12, doi: 10.18653/v1/d19-6001.

[44] R. Patel and E. Pavlick, ''Mapping language models to grounded conceptual spaces,'' in *Proc. 10th Int. Conf. Learn. Represent.*, 2022.

[45] Z. Li, Q. Xu, D. Zhang, H. Song, Y. Cai, Q. Qi, R. Zhou, J. Pan, Z. Li, V. Tu, Z. Huang, and T. Wang, ''GroundingGPT: Language enhanced multimodal grounding model,'' in *Proc. 62nd Annu. Meeting Assoc. Comput. Linguistics*, 2024, pp. 6657–6678, doi: 10.18653/v1/2024.acl-long.360.

[46] K. M. Collins, C. Wong, J. Feng, M. Wei, and J. B. Tenenbaum, ''Structured, flexible, and robust: Benchmarking and improving large language models towards more human-like behavior in out-of-distribution reasoning tasks,'' in *Proc. Annu. Meeting Cogn. Sci. Soc.*, 2022, pp. 869–875.

[47] W. Gurnee and M. Tegmark, ''Language models represent space and time,'' in *Proc. 12th Int. Conf. Learn. Represent.*, 2024.

[48] S. Trott and B. Bergen, ''Contextualized sensorimotor norms: Multidimensional measures of sensorimotor strength for ambiguous English words, in context,'' 2022, *arXiv:2203.05648*.

[49] A. Miklashevsky, ''Perceptual experience norms for 506 Russian nouns: Modality rating, spatial localization, manipulability, imageability and other variables,'' *J. Psycholinguistic Res.*, vol. 47, no. 3, pp. 641–661, Jun. 2018, doi: 10.1007/s10936-017-9548-1.

[50] L. J. Speed and A. Majid, ''Dutch modality exclusivity norms: Simulating perceptual modality in space,'' *Behav. Res. Methods*, vol. 49, no. 6, pp. 2204–2218, Dec. 2017, doi: 10.3758/s13428-017-0852-3.

[51] L. J. Speed and M. Brybaert, ''Dutch sensory modality norms,'' *Behav. Res. Methods*, vol. 54, no. 3, pp. 1306–1318, Jun. 2022, doi: 10.3758/s13428-021-01656-9.

[52] A. Vergallito, M. A. Petilli, and M. Marelli, ''Perceptual modality norms for 1,121 Italian words: A comparison with concreteness and imageability scores and an analysis of their impact in word processing tasks,'' *Behav. Res. Methods*, vol. 52, no. 4, pp. 1599–1616, Aug. 2020, doi: 10.3758/s13428-019-01337-8.

[53] G. Chedid, S. M. Brambati, C. Bedetti, A. E. Rey, M. A. Wilson, and G. T. Vallet, ''Visual and auditory perceptual strength norms for 3,596 French nouns and their relationship with other psycholinguistic variables,'' *Behav. Res. Methods*, vol. 51, no. 5, pp. 2094–2105, Oct. 2019, doi: 10.3758/s13428-019-01254-w.

[54] A. Miceli, E. Wauthia, L. Lefebvre, L. Ris, and I. S. Loureiro, ''Perceptual and interoceptive strength norms for 270 French words,'' *Frontiers Psychol.*, vol. 12, Jun. 2021, doi: 10.3389/fpsyg.2021.667271.

[55] I.-H. Chen, Q. Zhao, Y. Long, Q. Lu, and C.-R. Huang, ''Mandarin Chinese modality exclusivity norms,'' *PLoS ONE*, vol. 14, no. 2, Feb. 2019, Art. no. e0211336, doi: 10.1371/journal.pone.0211336.

[56] J. Lee and J.-A. Shin, ''The cross-linguistic comparison of perceptual strength norms for Korean, English and L2 English,'' *Frontiers Psychol.*, vol. 14, Jul. 2023, Art. no. 1188909, doi: 10.3389/fpsyg.2023.1188909.

[57] F. Günther, C. Dudschig, and B. Kaup, ''Symbol grounding without direct experience: Do words inherit sensorimotor activation from purely linguistic context?'' *Cognit. Sci.*, vol. 42, no. S2, pp. 336–374, May 2018, doi: 10.1111/cogs.12549.

[58] M. Louwerse and L. Connell, ''A taste of words: Linguistic context and perceptual simulation predict the modality of words,'' *Cognit. Sci.*, vol. 35, no. 2, pp. 381–398, Mar. 2011, doi: 10.1111/j.1551-6709.2010.01157.x.

[59] Meta AI. (2024). *Llama 3.2: Revolutionizing Edge AI and Vision With Open, Customizable Models*. [Online]. Available: https://ai.meta.com/blog/llama-3-2-connect-2024-vision-edge-mobile-devices/

[60] Meta AI. (2025). *The Llama 4 Herd: The Beginning of a New Era of Natively Multimodal AI Innovation*. [Online]. Available: https://ai.meta.com/blog/llama-4-multimodal-intelligence/

[61] S. Bai et al., ''Qwen2.5-VL technical report,'' 2025, *arXiv:2502.13923*.

[62] S. Shahriar, B. D. Lund, N. R. Mannuru, M. A. Arshad, K. Hayawi, R. V. K. Bevara, A. Mannuru, and L. Batool, ''Putting GPT-4o to the sword: A comprehensive evaluation of language, vision, speech, and multimodal proficiency,'' *Appl. Sci.*, vol. 14, no. 17, p. 7782, Sep. 2024, doi: 10.3390/app14177782.

[63] A. B. Abacha, W.-W. Yim, Y. Fu, Z. Sun, M. Yetisgen, F. Xia, and T. Lin, ''MEDEC: A benchmark for medical error detection and correction in clinical notes,'' 2024, *arXiv:2412.19260*.

[64] H. D. Critchley and N. A. Harrison, ''Visceral influences on brain and behavior,'' *Neuron*, vol. 77, no. 4, pp. 624–638, Feb. 2013, doi: 10.1016/j.neuron.2013.02.008.

[65] L. Connell, D. Lynott, and B. Banks, ''Interoception: The forgotten modality in perceptual grounding of abstract and concrete concepts,'' *Phil. Trans. Roy. Soc. B, Biol. Sci.*, vol. 373, no. 1752, Aug. 2018, Art. no. 20170143, doi: 10.1098/rstb.2017.0143.

[66] G. Dove, L. Barca, L. Tummolini, and A. M. Borghi, ''Words have a weight: Language as a source of inner grounding and flexibility in abstract concepts,'' *Psychol. Res.*, vol. 86, no. 8, pp. 2451–2467, Nov. 2022, doi: 10.1007/s00426-020-01438-6.

[67] P. Virtanen et al., ''SciPy 1.0: Fundamental algorithms for scientific computing in Python,'' *Nature Methods*, vol. 17, no. 3, pp. 261–272, Mar. 2020, doi: 10.1038/s41592-019-0686-2.

[68] B. Diedenhofen and J. Musch, ''Cocor: A comprehensive solution for the statistical comparison of correlations,'' *PLoS ONE*, vol. 10, no. 4, Apr. 2015, Art. no. e0121945, doi: 10.1371/journal.pone.0121945.

[69] N. C. Silver and W. P. Dunlap, ''Averaging correlation coefficients: Should Fisher's z transformation be used?'' *J. Appl. Psychol.*, vol. 72, no. 1, pp. 146–148, 1987.

[70] A. Kuznetsova, P. B. Brockhoff, and R. H. B. Christensen, ''LmerTest package: Tests in linear mixed effects models,'' *J. Stat. Softw.*, vol. 82, no. 13, 2017, doi: 10.18637/jss.v082.i13.

[71] D. Bates, M. Mächler, B. Bolker, and S. Walker, ''Fitting linear mixed-effects models Usinglme4,'' *J. Stat. Softw.*, vol. 67, no. 1, pp. 1–48, 2015, doi: 10.18637/jss.v067.i01.

[72] R. Speer, ''rspeer/wordfreq: V3.0,'' Tech. Rep., 2022, doi: 10.5281/zenodo.7199437.

[73] K. Erk, ''What do you know about an alligator when you know the company it keeps?'' *Semantics Pragmatics*, vol. 9, pp. 1–63, Apr. 2016, doi: 10.3765/sp.9.17.

[74] E. M. Buchanan, K. D. Valentine, and N. P. Maxwell, ''English semantic feature production norms: An extended database of 4436 concepts,'' *Behav. Res. Methods*, vol. 51, no. 4, pp. 1849–1863, Aug. 2019, doi: 10.3758/s13428-019-01243-z.

[75] C. Xiao, J. Ye, R. M. Esteves, and C. Rong, ''Using Spearman's correlation coefficients for exploratory data analysis on big dataset,'' *Concurrency Comput., Pract. Exper.*, vol. 28, no. 14, pp. 3866–3878, Sep. 2016, doi: 10.1002/cpe.3745.

[76] Y. Wang, X. Ma, G. Zhang, Y. Ni, A. Chandra, S. Guo, W. Ren, A. Arulraj, X. He, Z. Jiang, T. Li, M. Ku, K. Wang, A. Zhuang, R. Fan, X. Yue, and W. Chen, ''MMLU-Pro: A more robust and challenging multi-task language understanding benchmark,'' in *Proc. Adv. Neural Inf. Process. Syst.*, vol. 37, 2024, pp. 95266–95290.

[77] X. Yue et al., ''MMMU: A massive multi-discipline multimodal understanding and reasoning benchmark for expert AGI,'' in *Proc. IEEE/CVF Conf. Comput. Vis. Pattern Recognit. (CVPR)*, Jun. 2024, pp. 9556–9567.

[78] Novita AI. (Feb. 4, 2025). *Llama 3.3 70B Vs Llama 3.2 90B: Text Mastery or Visual Power*. [Online]. Available: https://blogs.novita.ai/llama-3-3-70b-vs-llama-3-2-90b/

[79] A. Koukounas, G. Mastrapas, S. Eslami, B. Wang, M. Kalim Akram, M. Günther, I. Mohr, S. Sturua, N. Wang, and H. Xiao, ''Jina-clip-v2: Multilingual multimodal embeddings for text and images,'' 2024, *arXiv:2412.08802*.

[80] Z. Gong, C. Mai, and Y. Huang, ''AsCL: An asymmetry-sensitive contrastive learning method for image-text retrieval with cross-modal fusion,'' in *Proc. IEEE Int. Conf. Multimedia Expo (ICME)*, Jul. 2024, pp. 1–6.

[81] Z. Zhou, X. Ning, K. Hong, T. Fu, J. Xu, S. Li, Y. Lou, L. Wang, Z. Yuan, X. Li, S. Yan, G. Dai, X.-P. Zhang, Y. Dong, and Y. Wang, ''A survey on efficient inference for large language models,'' 2024, *arXiv:2404.14294*.

[82] N. Sardana, J. Portes, S. Doubov, and J. Frankle, ''Beyond Chinchilla-optimal: Accounting for inference in language model scaling laws,'' 2023, *arXiv:2401.00448*.

[83] Y. Liu, W. Chen, Y. Bai, X. Liang, G. Li, W. Gao, and L. Lin, "Aligning cyber space with physical world: A comprehensive survey on embodied AI," *IEEE/ASME Trans. Mechatronics*, early access, Jul. 28, 2025, doi: 10.1109/TMECH.2025.3574943.

[84] Z. Feng, R. Xue, L. Yuan, Y. Yu, N. Ding, M. Liu, B. Gao, J. Sun, X. Zheng, and G. Wang, "Multi-agent embodied AI: Advances and future directions," 2025, *arXiv:2505.05108*.

[85] R. Mon-Williams, G. Li, R. Long, W. Du, and C. G. Lucas, "Embodied large language models enable robots to complete complex tasks in unpredictable environments," *Nature Mach. Intell.*, vol. 7, no. 4, pp. 1–10, Mar. 2025.

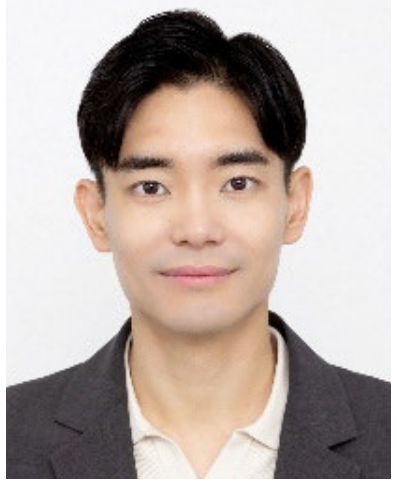

**JONGHYUN LEE** received the Ph.D. degree in English linguistics from Seoul National University, Seoul, South Korea. He is currently an Assistant Professor with the Division of Global Studies (English Studies Major), Korea University, Sejong Campus, Sejong, South Korea. His research interests include psycholinguistics, neurolinguistics, and computational linguistics.

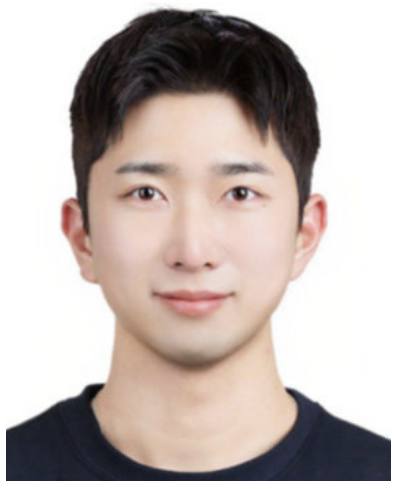

**DOJUN PARK** received the B.Tech. degree in human ICT convergence and the B.A. degree in English language and literature from Konkuk University, Seoul, South Korea, in 2021, the M.Sc. degree in computational linguistics from the University of Stuttgart, Germany, in 2023. He is currently an Intern with the Artificial Intelligence Institute of Seoul National University (AIIS), actively engaged in a project focusing on the human-level evaluation of LLMs. His research interests include the development of generative models, such as machine translation and LLMs, and their evaluation.

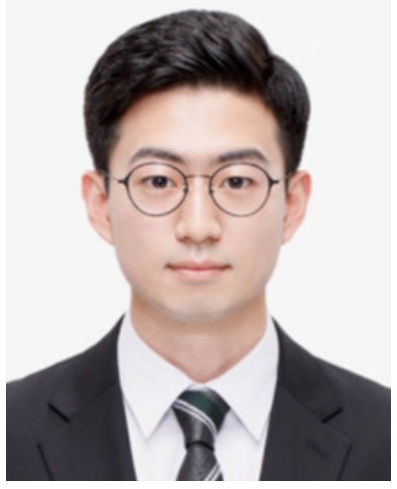

**JIWOO LEE** received the B.A. degree in German language and literature from Kyonggi University, Suwon, South Korea, and the M.A. degree in German language and literature from the Department of German Language and Literature, Seoul National University, Seoul, South Korea, in 2023, where he is currently pursuing the Ph.D. degree. He is a Researcher with the Artificial Intelligence Institute of Seoul National University (AIIS). His research interests include pragmatics in general and integrating those aspects with LLMs.

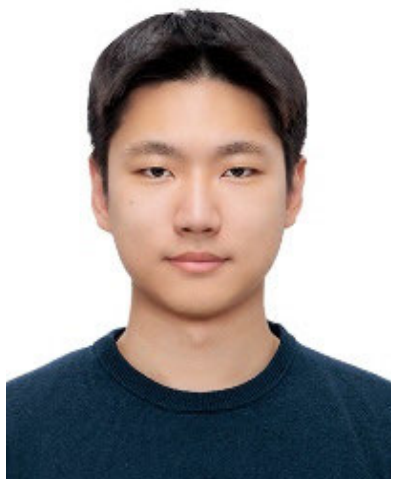

**HOEKEON CHOI** received the B.A. degree in English education from Incheon National University, Incheon, South Korea, in 2022. He is currently pursuing the M.A degree with the Department of English Language and Literature, Seoul National University, Seoul, South Korea. He is a Researcher with the Artificial Intelligence Institute of Seoul National University (AIIS). His research interests include second language acquisition, psycho-neurolinguistics, embodied cognition.

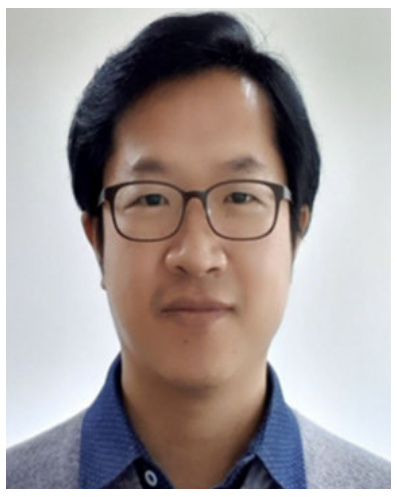

**SUNG-EUN LEE** is currently a Professor with the Department of Interdisciplinary Program in Cognitive Science and the Department of German Language and Literature, and the Director of the Center for Language Cognition and Artificial Intelligence at Seoul National University. His main research interests include neurolinguistics, and he studies language processing in the human brain through experimental studies using brain imaging with the Brain and Humanities Laboratory (BHL). He is also interested in combining humanities research topics with fields such as engineering, psychology, and medicine, and is actively involved in interdisciplinary research that combines artificial intelligence technologies with cognitive neuroscience methodologies.